\documentclass[letterpaper, 10 pt, journal, twoside]{IEEEtran}
\usepackage{cite}
\usepackage{amsmath,amssymb,amsfonts}
\usepackage[T1]{fontenc}
\usepackage{hyperref}
\usepackage{graphicx}
\usepackage{booktabs}
\usepackage{threeparttable}
\usepackage{multirow}
\usepackage{textcomp}
\usepackage{xcolor}
\usepackage[ruled,linesnumbered]{algorithm2e}
\usepackage{lipsum}
\usepackage{makecell}
\usepackage[resetlabels]{multibib}

\newcites{supp}{References for Supplementary Material}

\makeatletter 

\newcommand{\makesupplementtitle}{%
  \clearpage 
  \twocolumn[%
    \begin{center}
      {\Huge \@title \par} 
      \vspace{2.0em} 
    \end{center}
  ]%
}

\makeatother

\hypersetup{hidelinks}

\begin{document}

\bstctlcite{IEEEexample:BSTcontrol}

\markboth{ IEEE ROBOTICS AND AUTOMATION LETTERS. PREPRINT VERSION. ACCEPTED JANUARY, 2026}
{Chen \MakeLowercase{\textit{et al.}}: Accurate Calibration and Robust LiDAR-Inertial Odometry for Spinning Actuated LiDAR Systems}

\author{Zijie Chen, Xiaowei Liu, Yong Xu, \IEEEmembership{Member, IEEE}, \\Shenghai Yuan, \IEEEmembership{Member, IEEE}, Jianping Li, \IEEEmembership{Member, IEEE}, and Lihua Xie, \IEEEmembership{Fellow, IEEE}
\thanks{Manuscript received: September 14, 2025.; Revised: November 10, 2025; Accepted: January 12, 2026.}
\thanks{This paper was recommended for publication by Editor Lucia Pallottino upon evaluation of the Associate Editor and Reviewers' comments. This work was supported in part by the National Natural Science Foundation of China under Grants (62525303, 62121004, U22A2044, 62576113, 62576114, 62236001). (Corresponding author: Yong Xu.)}
\thanks{Zijie Chen, Xiaowei Liu, and Yong Xu are with Engineering Research Center of Low-Altitude Perception and Detection Technology \& Intelligent IoT, Ministry of Education, School of Automation, Guangdong University of Technology, Guangzhou 510006, China (email: zijiechenrobotics@foxmail.com; xiaoweiliurobot@foxmail.com; yxu@gdut.edu.cn).}
\thanks{Shenghai Yuan, Jianping Li, and Lihua Xie are with the School of Electrical and Electronic Engineering, Nanyang Technological University 63979, Singapore (email: shyuan@ntu.edu.sg; jianping.li@ntu.edu.sg; elhxie@ntu.edu.sg).}
\thanks{Digital Object Identifier (DOI): see top of this page.}
}

\title{Accurate Calibration and Robust LiDAR-Inertial Odometry for Spinning Actuated LiDAR Systems}



\maketitle



\begin{abstract}
Accurate calibration and robust localization are fundamental for downstream tasks in spinning actuated LiDAR applications. Existing methods, however, require parameterizing extrinsic parameters based on different mounting configurations, limiting their generalizability. Additionally, spinning actuated LiDAR inevitably scans featureless regions,  which complicates the balance between scanning coverage and localization robustness. To address these challenges, this letter presents a targetless LiDAR-motor calibration (LM-Calibr) on the basis of the Denavit-Hartenberg convention and an environmental adaptive LiDAR-inertial odometry (EVA-LIO). LM-Calibr supports calibration of LiDAR-motor systems with various mounting configurations. Extensive experiments demonstrate its accuracy and convergence across different scenarios, mounting angles, and initial values. Additionally, EVA-LIO adaptively selects downsample rates and map resolutions according to spatial scale. This adaptivity enables the actuator to operate at maximum speed, thereby enhancing scanning completeness while ensuring robust localization, even when LiDAR briefly scans featureless areas. The source code and hardware design are available on GitHub: \textcolor{blue}{\href{https://github.com/zijiechenrobotics/lm_calibr}{github.com/zijiechenrobotics/lm\_calibr}}. The video is available at \textcolor{blue}{\href{https://youtu.be/cZyyrkmeoSk}{youtu.be/cZyyrkmeoSk}}
\end{abstract}


%
\IEEEpeerreviewmaketitle

\section{Introduction and Related Work}

\IEEEPARstart{S}{pinning} actuated LiDAR (LiDAR-motor) systems, characterized by their omnidirectional field of view (FOV) and high resolution, are well-suited for robotic navigation \cite{chen_nan} and mapping \cite{LOAM, rotating_lidar_uav,  rls_lcd, or_lim, rss_liwom}. Accurate calibration and robust LiDAR-inertial odometry (LIO) serve as fundamental prerequisites for these downstream tasks. However, both aspects remain challenges to date.

Existing LiDAR-motor calibration methods \cite{iros_2015_calib, FGRSC, limo_calib, 2d_lrf, calib_multi_line, calib_2d, spatial_temporal_calib, boresight_calib} predominantly focus on targetless approaches, but they lack a general parameterization of the extrinsic parameters. In \cite{iros_2015_calib}, a self-calibration method leverages the intrinsic redundancy of recorded point clouds to achieve reliable calibration without the need for specific targets. Building upon it, FGRSC enhances robustness in unprepared scenes by employing a filtered grid strategy with randomized sampling, ensuring reliable extraction of planar features \cite{FGRSC}. To further improve computational efficiency, LiMo-Calib introduces a feature selection strategy coupled with a reweighting mechanism to accelerate the optimization of planar constraints \cite{limo_calib}. Despite these advancements, existing methods are limited by the observability of the extrinsic parameters. Early works claim that unidirectional LiDAR-motor systems have only four observable degrees of freedom (DOF) \cite{auto_calib, full_dof}. Specifically, the rotational and translational components parallel to the motor's spinning axis are unobservable. For instance, FGRSC calibrates only yaw and pitch. When the yaw is $\pm 90^\circ$, the LiDAR’s y-axis aligns with the spinning axis, making the pitch unobservable. Similarly, LiMo-Calib estimates only roll and pitch. If the pitch is $\pm 90^\circ$, the LiDAR’s x-axis aligns with the spinning axis, which results in the roll being unobservable.  These shortcomings underscore the importance of developing a general LiDAR-motor calibration method.

\begin{figure*}[t]
\centering
\includegraphics[width=0.98\textwidth]{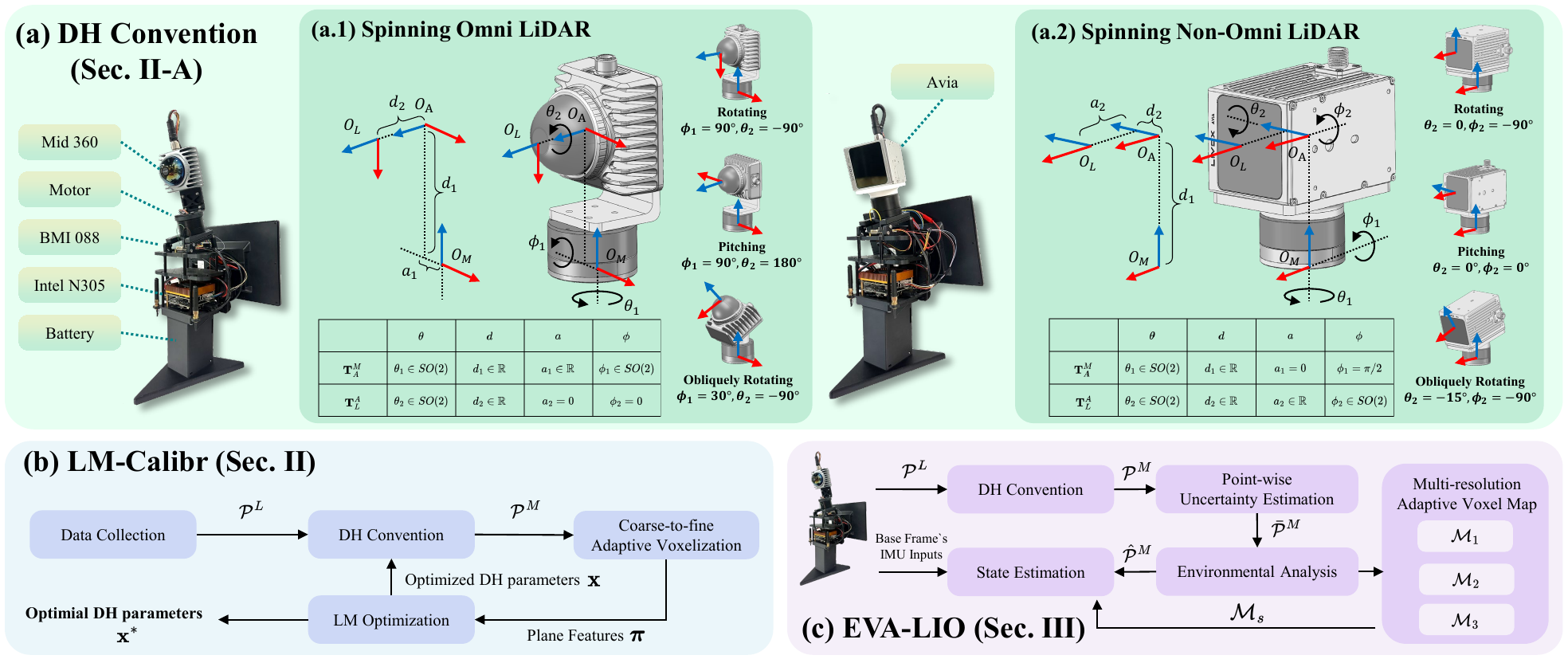}
\caption{(a.1) and (a.2) show the DH convention of spinning omni and non-omni LiDARs. (b) and (c) present the block diagrams of LM-Calibr and EVA-LIO.}
\label{fig:framework}
\end{figure*}

When deployed in LIOs\cite{IN2LAAMA, steam_icp, efficient_continuous}, spinning LiDAR inevitably scans featureless regions in challenging environments, such as long corridors or stairwells. These scenarios create a trade-off between scanning coverage and localization robustness. To address this challenge, Zebedee augments 2D LiDAR coverage by mounting the sensor on a moving platform via a passive linkage mechanism \cite{zebedee}. However, this coverage-enhancement strategy neglects the odometry observability. In contrast,  $\alpha$LiDAR introduces an active rotational mechanism that dynamically adjusts the FOV to focus on specific areas of interest  \cite{alpha_lidar}. Nevertheless, this selective scanning compromises the perception of environmental changes and mapping consistency. As opposed to adjusting the FOV, UA-MPC proposes an uncertainty-aware motor speed strategy \cite{ua_mpc}. It predicts the odometry observability and dynamically adjusts the motor speed. Its flaw is that the motor speed must continuously adapt to environmental changes. In some scenarios\footnote{At 1:30 of the  \href{https://www.youtube.com/watch?v=zkbm0Tkp-PM&t=90s}{https://www.youtube.com/watch?v=zkbm0Tkp-PM\&t=90s}}
, the motor speed difference between consecutive frames exceeds 10 rad/s. Such rapid variation not only impacts the dynamic balancing of the sensing systems but also increases energy consumption. Furthermore, due to physical constraints, the actuator cannot instantaneously reorient the LiDAR’s FOV toward feature-rich areas. In a long corridor, spinning Avia continuous coverage of the feature-rich areas would require a rotational speed that exceeds the operational limits of practical actuators. Therefore, it is worthwhile to explore an odometry that strikes a balance between scanning coverage and localization robustness.

To address the challenges discussed above, this letter presents a general LiDAR-motor calibration method and an environmental adaptive localization approach for the spinning actuated LiDAR systems. The main contributions are as follows.

\begin{itemize}
\item A unified kinematic model for the LiDAR-motor systems is introduced based on the Denavit–Hartenberg (DH) convention. Compared with the previous parameterization \cite{iros_2015_calib, FGRSC, limo_calib, 2d_lrf, calib_multi_line, calib_2d, spatial_temporal_calib, boresight_calib}, this formulation is applicable to various LiDAR-motor configurations, including rotating, pitching, and obliquely rotating, as illustrated in Fig.~\ref{fig:framework}(a.1) and (a.2).
\item A targetless LiDAR-motor calibration (LM-Calibr), built upon the DH convention, is proposed. Extensive simulation and real-world experiments demonstrate that this method does not require special target objects or structured environments with flat surfaces. It achieves accuracy and convergence under different scenarios, mounting angles, and initial parameters (see Section \ref{sec:lm_calibr_exp}).
\item An environmental adaptive LiDAR-inertial odometry (EVO-LIO) is developed. EVA-LIO enables the actuator to operate at maximum speed for improved scanning coverage and maintains robust localization in various challenging scenarios (see Section \ref{sec:eva_lio_exp}).
\end{itemize}

\section{LiDAR-Motor Calibration}\label{sec:LM_calibr}

The unidirectional LiDAR-motor systems discussed in this work are shown in Fig.~\ref{fig:framework}(a). The motor actuates the LiDAR to increase the system’s FOV. Due to the mechanical structure of the system, the center of the actuator $\{ M \}$ may not align with the LiDAR rotating mirror $\{ L \}$, resulting in a rigid-body transformation $\mathbf{T}^M_L\in SE(3)$. Building upon this geometric model, this section presents a parameterization method for $\mathbf{T}^M_L$ and proposes a corresponding calibration procedure.

\subsection{Denavit-Hartenberg Convention}\label{sec:DH_convention}

The unidirectional LiDAR-motor system can be regarded as an arm with 1-DOF. Specifically, the motor frame $\{ M \}$ is equivalent to the base of the arm, and the LiDAR frame $\{ L \}$ is comparable to the end effector. As shown in Fig.~\ref{fig:framework}(a), the transformation between $\{ M \}$ and $\{ L \}$ can be formulated by the DH parameters
\begin{equation*}
\mathcal{DH} \triangleq \begin{bmatrix}
\theta_1 & d_1 & a_1 & \phi_1 & \theta_2 & d_2 & a_2 & \phi_2
\end{bmatrix}.
\end{equation*}

The laser point in $\{ M \}$ is given by

\begin{align}\label{eq:full_rigid_trans}
\mathbf{p}^M &= \mathbf{T}^M_L \mathbf{p}^L \\
&= \mathbf{R}_z(\theta_1)( \mathbf{R}_x(\phi_1)\mathbf{R}_z(\theta_2)(\mathbf{R}_x(\phi_2)\mathbf{p}^L + \mathbf{t}_1 ) + \mathbf{t}_2 ), \nonumber
\end{align}
where
\begin{equation*}
\begin{split}
\mathbf{R}_x(\phi) &= \begin{bmatrix}
1 & 0 & 0\\
0 & \cos(\phi) & -\sin(\phi) \\
0 & \sin(\phi) & \cos(\phi)
\end{bmatrix}, \mathbf{t}_1 = \begin{bmatrix}
a_2 \\ 
0 \\
d_2
\end{bmatrix},\\
\mathbf{R}_z(\theta) &= \begin{bmatrix}
\cos(\theta) & -\sin(\theta) & 0 \\
\sin(\theta) & \cos(\theta) & 0 \\
0 & 0 & 1
\end{bmatrix}, \mathbf{t}_2 = \begin{bmatrix}
a_1 \\ 
0 \\
d_1
\end{bmatrix}.
\end{split}
\end{equation*}

The DH convention consists of eight parameters. The encoder measures the angle $\theta_1$, and the mechanical drawing provides the translation $d_1$, while the remaining parameters require calibration. Note that the encoder and translation $d_1$ installation errors can be compensated during downstream calibration, e.g., the LiDAR-IMU calibration of LIO.

As mentioned in the Introduction, the rotational and translational components parallel to the motor’s spinning axis are unobservable \cite{auto_calib, full_dof}. To avoid this unobservability, LM-Calibr introduces two sets of DH parameters to represent the extrinsic parameters of spinning omni LiDARs (the rotating mirror rotates about the z-axis) and non-omni LiDARs (the rotating mirror rotates around the x-axis).  For the spinning omni LiDAR, the constant parameters are $a_2 = 0$ and $\phi_2 = 0$, while the unknown parameters are $\theta_2, d_2, a_1$, and $\phi_1$. The unobservability case occurs only when $\phi_1 \approx 0$ or $\pi$, which corresponds to the scenario where the omni lidar’s rotating mirror is parallel to the motor's spinning axis. Such a mounting configuration is physically meaningless for increasing FOV. In terms of the spinning non-omni LiDAR, the constant parameters are $\alpha_1 = 0$ and $\phi_1 = \pi/2$. The parameters to be calibrated are $\theta_2, d_2, a_2$, and $\phi_2$. The calibration degradation only occurs for the case $\theta_2 \approx \pm \pi$, which corresponds to the non-omni LiDAR's rotating mirror being parallel to the motor's rotation axis.

In view of the observability analysis in Section C of the supplementary material,  the DH convention allows for the representation of any mounting configuration of the LiDAR-motor systems, demonstrating both versatility and portability. Fig.~\ref{fig:framework}(a.1) and (a.2) show some typical mounting for the LiDAR-motor systems, i.e., rotating, pitching, and obliquely rotating. To facilitate the subsequent description, the parameters that need to be calibrated are defined as $\mathbf{x} \triangleq  \begin{bmatrix}
\bar{\theta} & \bar{d} & \bar{a} & \bar{\phi}
\end{bmatrix}^\top$, and the transformation \eqref{eq:full_rigid_trans} can be rewritten as
\begin{equation}\label{eq:simple_rigid_trans}
\mathbf{p}^M(\mathbf{x}) = \mathbf{T}^M_L(\mathbf{x})\mathbf{p}^L.
\end{equation}

\subsection{Denavit-Hartenberg Parameters Calibration}\label{sec:DH_calib}

If the DH parameters are incorrect, the transformed point cloud exhibits significant distortion. Leveraging this phenomenon, LM-Calibr minimizes the thickness of the uncalibrated point cloud to obtain correct DH parameters \cite{balm_1}.  Specifically, the sensing system first remains stationary for a period to collect an uncalibrated point cloud $\mathcal{P}_L$, which is subsequently transformed from $\{ L \}$ to $\{ M \}$ via the transformation \eqref{eq:simple_rigid_trans}. The adaptive voxelization \cite{balm_1} is then applied to the transformed point cloud $\mathcal{P}_M$ to extract the plane features $\boldsymbol{\pi} = (\pi_1,...,\pi_{M_f} )$. Finally, the Levenberg-Marquardt (LM) method is employed to minimize the thickness of each plane feature. The cost function is defined as
\begin{equation}\label{eq:lm_calibr_cost_func}
\min_{\mathbf{x}}(\sum^{M_f}_{i=1}\pi_{i}(\mathbf{x})),
\end{equation}
where
\begin{equation*}
\begin{split}
\pi_i(\mathbf{x}) &= \lambda_{\min}(\mathbf{A}_i(\mathbf{x})), \\
\mathbf{A}_i(\mathbf{x}) &= \frac{1}{N_i}\sum^{N_i}_{j=1}(\mathbf{p}^M_{ij}(\mathbf{x}) - \mathbf{q}_i(\mathbf{x}) ) ( \mathbf{p}^M_{ij}(\mathbf{x}) - \mathbf{q}_i(\mathbf{x}) )^\top, \\
\mathbf{q}_i (\mathbf{x}) &= \frac{1}{N_i}\sum^{N_i}_{j=1}\mathbf{p}^M_{ij}(\mathbf{x}), \mathbf{p}^M_{ij}(\mathbf{x}) = \mathbf{T}^M_L(\mathbf{x}) \mathbf{p}^L_{ij}.\\ 
\end{split}
\end{equation*}
The planar feature $\pi_i$ contains $N_i$ laser points. Its centroid and covariance matrix are denoted by $\mathbf{q}_i$ and $\mathbf{A}_i$. $\mathbf{p}^L_{ij}$ represents the $j$-th laser point in the $i$-th plane feature. $\lambda_{\min}(\mathbf{A})$ denotes the minimum eigenvalue of matrix $\mathbf{A}$. The derivative of the cost function \eqref{eq:lm_calibr_cost_func} is described in Section A of the supplementary material.

The complete calibration process is summarized in Fig.~\ref{fig:framework}(b). To ensure robustness in challenging scenarios and poor initial parameters, LM-Calibr employs a coarse-to-fine strategy to adjust the root voxel size of the adaptive voxelization depending on the number of iterations. During the first and second iterations, the root voxel size is set to be $1 m$. For the third and fourth iterations, the root voxel size is reduced to be $0.5 m$. The root voxel size for further iterations is set to be $0.25 m$.

\section{Environmental Adaptive LiDAR-Inertial Odometry}\label{sec:ada_lio}

EVA-LIO preserves maximum actuator speed to enhance scanning coverage \cite{ua_mpc}. The following aspects ensure its robustness in the challenging scenarios (e.g., long corridors and stairwells).

\begin{itemize}
\item \textbf{EVA-LIO tightly couples the base frame IMU and the laser points in $\{ M \}$.} Taking the LiDAR’s built-in IMU as LIO input may lead to saturation during aggressive motion, since self-motion combined with motor rotation may exceed the IMU’s measurement range. In contrast, utilizing the base frame IMU inputs prevents this situation and improves robustness. For example, when the scanning pattern momentarily covers the featureless areas, the base frame IMU provides better initial estimates in unobservable directions, reducing the risk of LIO degradation.
\item \textbf{EVA-LIO estimates the uncertainty of laser points.} Typically, LiDAR operates at sampling rates above $10 kHz$, while the encoder sampling rate is limited to approximately $1 kHz$. This mismatch impedes accurate interpolation of the beam angle for each laser point. To address this challenge, the measurement noise of each laser point is modeled, which reduces the influence of bearing direction uncertainty on localization accuracy (see Section B of the supplementary material).
\item \textbf{EVA-LIO adapts system parameters (i.e., downsample rate and map resolution) based on the environmental scale.} Although some LIO methods adopt adaptive voxel subdivision according to local geometric features \cite{adaptive_voxel, voxel_slam}, the root voxel size still requires manual adjustment for complex scenarios. In contrast, EVA-LIO achieves robust localization across different scenarios without manual tuning.
\end{itemize}


The main process of EVA-LIO is shown in Fig.~\ref{fig:framework}(c). First, the point cloud $\mathcal{P}^L$ is transformed to $\{ M \}$ via the DH convention. The transformed point cloud $\mathcal{P}^M$ is then processed through undistortion and uncertainty modeling to produce the processed point cloud $\bar{\mathcal{P}}^M$. The environmental analysis module evaluates the spatial scale and determines the downsample rate $V_{s}$ and the map $\mathcal{M}_{s}$. Each laser point in the downsampled point cloud $\hat{\mathcal{P}}^M$ is then transformed into the world frame to match the plane features in $\mathcal{M}_{s}$. Finally, the system state 
$\mathbf{x}_k \triangleq \begin{bmatrix}
\mathbf{R}^W_{B_k} & \mathbf{t}^W_{B_k} & \mathbf{v}^W_{B_k} & \mathbf{b}_{\alpha, k} & \mathbf{b}_{g, k}
\end{bmatrix}^\top$
is estimated using maximum a posteriori (MAP) estimation \cite{ig_lio}
\begin{equation}\label{eq:ada_lio_cost_func}
\min_{\mathbf{x}_k}\bigg\{ \sum_{i\in\hat{\mathcal{P}}^M_k}\left\|\mathbf{r}^{\mathcal{L}}_{i}\right\|^2_{\boldsymbol{\Omega}^{\mathcal{L}}_{i}} + \left\|\mathbf{r}_k^{\mathcal{I}}\right\|^2_{\boldsymbol{\Omega}_k^{\mathcal{I}}}\bigg\},
\end{equation}
where $\left\|\mathbf{r}\right\|^2_{\boldsymbol{\Omega}}=\mathbf{r}^\top\boldsymbol{\Omega}\mathbf{r}$. $\mathbf{R}^W_{B}\in SO(3), \mathbf{t}^W_{B}\in\mathbb{R}^3$, and $\mathbf{v}^W_{B}\in\mathbb{R}^3$ are the rotation, position, and velocity of the base frame $\{ B \}$ in the world frame $\{W\}$, respectively. $\mathbf{b}_{\alpha}\in\mathbb{R}^3$ and $\mathbf{b}_{g}\in\mathbb{R}^3$ denote the biases of a three-axis gyroscope and a three-axis accelerometer. The residual $\mathbf{r}^{\mathcal{L}}$ represents the point-to-plane distance \cite{adaptive_voxel} and $\mathbf{r}^{\mathcal{I}}$ is the IMU midpoint integration constraint \cite{ig_lio}. The matrices $\boldsymbol{\Omega}^{\mathcal{L}}$ and $\boldsymbol{\Omega}^{\mathcal{I}}$ are the corresponding information matrices.

\subsection{Environmental Analysis Module}

EVA-LIO manages multiple adaptive voxel maps \cite{adaptive_voxel} with different root voxel sizes, denoted as $\boldsymbol{\mathcal{M}}=(\mathcal{M}_1,\mathcal{M}_2,\mathcal{M}_3)$. For each incoming LiDAR frame, the environmental analysis module selects a downsample rate $V_s$ and a voxel map $\mathcal{M}_s$ according to the spatial scale, as shown in Algorithm \ref{alg:env_analysis}. Specifically, a downsampled panoramic map $\bar{\mathbf{M}}$ is first constructed by merging the processed point cloud  $\bar{\mathcal{P}}^M$ with a history point cloud $\mathcal{P}_h$. The historical frames retained in $\mathcal{P}_h$ depend on the motor speed (e.g., eight frames at $7.85 rad/s$). 
The spatial scale $s$ is then evaluated from $\bar{\mathbf{M}}$ by
\begin{equation}\label{eq:spatical_scale}
s = \frac{1}{|\bar{\mathbf{M}}|} \sum^{|\bar{\mathbf{M}}|}_{i = 1} ||\mathbf{p}^W_i - \mathbf{q}_{\bar{\mathbf{M}}}||_2 , 
\mathbf{q}_{\bar{\mathbf{M}}} = \frac{1}{|\bar{\mathbf{M}}|}\sum^{|\bar{\mathbf{M}}|}_{i = 1} \mathbf{p}^W_i, 
\end{equation}
 where $\mathbf{p}^W_i \in \bar{\mathbf{M}}$. The vector $\mathbf{q}_{\bar{\mathbf{M}}}$ is the centroid of the panoramic map $\bar{\mathbf{M}}$. The environmental type is then classified as narrow, normal, or wide based on the criteria $s_1$ and $s_2$, which are calibrated in corresponding typical scenarios. Depending on the classification, EVA-LIO utilizes the selected map $\mathcal{M}_s$ and downsampled LiDAR's scan $\hat{\mathcal{P}}^M$ for subsequent state estimation.

\begin{algorithm}[t]
\caption{Environmental Analysis Module}\label{alg:env_analysis}
\LinesNumbered 
\SetAlgoNoEnd
\KwIn{
Maps $\boldsymbol{\mathcal{M}}=(\mathcal{M}_1,\mathcal{M}_2,\mathcal{M}_3)$,
downsample rates $
\mathbf{V} = (V_1, V_2, V_3)$,
voxel size $V_e = 5m$,
narrow criteria $s_1$,
wide criteria $s_2$,
processed point cloud $\bar{\mathcal{P}}^M$,
history point cloud $\mathcal{P}_h$.
} 
\KwOut{Map $\mathcal{M}_s$, 
downsampled point cloud $\hat{\mathcal{P}}^M$.} 

Panoramic map $\mathbf{M} = \bar{\mathcal{P}}^M \cup  \mathcal{P}_h$\;

Downsample $\mathbf{M}$ with $V_e \to \bar{\mathbf{M}}$\;

Evaluate the spatial scale $d$ of $\bar{\mathbf{M}}$ via \eqref{eq:spatical_scale}\;

\If{$s > s_2$ }{
$V_s = V_3, \mathcal{M}_s = \mathcal{M}_3$; \tcp*[f]{Wide}
}
\ElseIf{$s < s_1$}{
$V_s = V_1, \mathcal{M}_s = \mathcal{M}_1$; \tcp*[f]{Narrow}
}
\Else {
$V_s = V_2, \mathcal{M}_s = \mathcal{M}_2$; \tcp*[f]{Normal}
}
\textbf{end if}

Downsample $\bar{\mathcal{P}}^M$ with $V_s \to \hat{\mathcal{P}}^M$\;
\end{algorithm}

\section{Experiments}\label{sec:experiments}

This section presents both simulation and real-world experiments to validate the proposed LM-Calibr and EVA-LIO. To generate synthetic data, MARSIM \cite{marsim} is employed to simulate the spinning Mid360 and Avia. The synthetic IMU data are generated from ground-truth (GT) trajectories. The simulation parameters are listed in Table \ref{table:sim_param}. 
All experiments are conducted with multi-threading on an i3-N305 CPU (3.0 GHz, 16GB). The attached video \footnote{\href{https://youtu.be/cZyyrkmeoSk}{https://youtu.be/cZyyrkmeoSk}} details the experimental process.

\begin{table}[t]
\caption{Simulation Parameters and Distribution of Ground-Truth DH Parameters}\label{table:sim_param}
\setlength{\tabcolsep}{0.4mm}
\centering
\scriptsize
\begin{threeparttable}
\begin{tabular}{cc|cc}
\toprule[1pt]
Parameter                  & Value                          & Parameter                & Value                        \\ \hline
Mid360 Range (m) / Density & 40 / 200k                      & Accel. Random Walk       & 3e-03                        \\
Mid360 FOV (H$\times$V)   & $360^\circ \times 59^\circ$    & $\bar{d}$                & $\mathcal{U}(-0.1, 0.1)$     \\
Avia Range (m) / Density   & 100 / 240k                     & $\bar{a}$                & $\mathcal{U}(-0.1, 0.1)$     \\
Avia FOV (H$\times$V)     & $70.4^\circ \times 77.2^\circ$ & $\bar{\theta}$ of Mid360 & $\mathcal{U}(-\pi, \pi)$     \\
Encoder Freq. (Hz)         & 200                            & $\bar{\theta}$ of Avia   & $\mathcal{U}(-\pi/8, \pi/8)$ \\
IMU Freq. (Hz)             & 200                            & $\bar{\phi}$ of Mid360   & $\mathcal{U}(0, \pi)$        \\
Gyro. White Noise          & 1.7e-04                        & $\bar{\phi}$ of Avia     & $\mathcal{U}(-\pi, \pi)$     \\
Gyro. Random Walk          & 2e-05                          & Trans. Noise (m)         & 0.05                         \\
Accel. White Noise         & 2e-03                          & Rot. Noise (deg)         & 5   \\ \bottomrule[1pt] 
\end{tabular}
\end{threeparttable}
\end{table}

\subsection{The Generalizability and Accuracy of LM-Calibr} \label{sec:lm_calibr_exp}

This subsection verifies the generalizability of LM-Calibr through Monte Carlo simulations and subsequently compares the accuracy of LM-Calibr and LiMo-Calib in simulated and real-world scenarios. To guarantee simulation realism, NTU \cite{mcd}, Cave \cite{super_loc}, and Botanic Garden (BG) \cite{botanic_garden} are adopted as MARSIM input. All methods terminate the optimization if the cost change is below 1e-6.

\begin{figure}[t]
\centering
\includegraphics[width=0.48\textwidth]{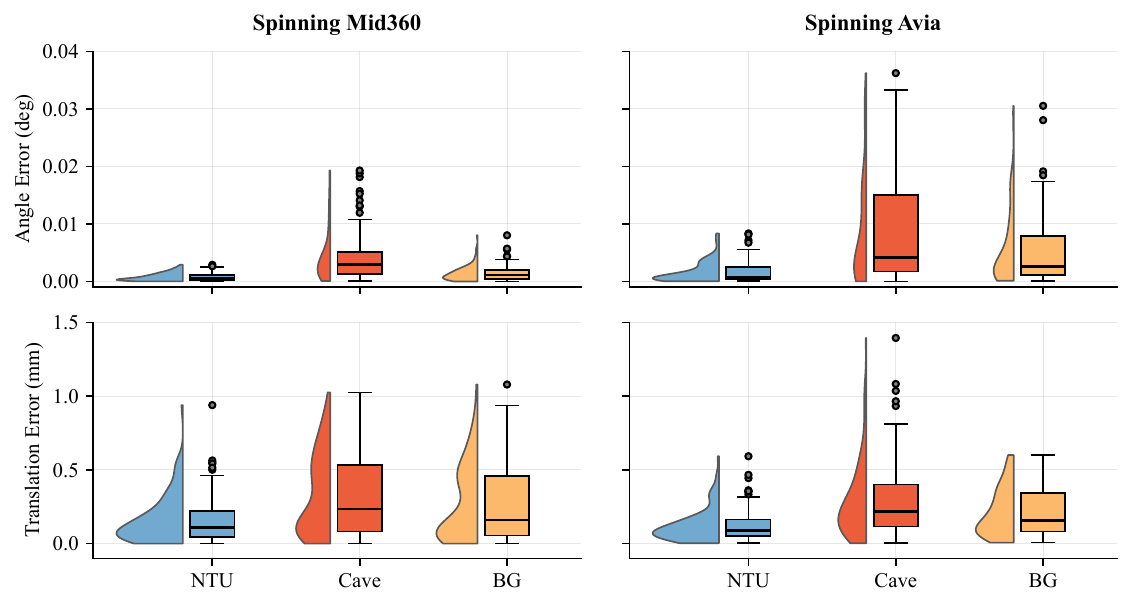}
\caption{The distribution of calibration errors in the Monte-Carlo simulations.}
\label{fig:monte_carlo_calib}
\end{figure}

\subsubsection{Generalizability Verification}

The Monte-Carlo simulations for LM-Calibr are conducted under various scenarios and mounting configurations. LiMo-Calibr is excluded here due to degeneracy in certain cases (e.g., the "Rotating" configuration in Fig.~\ref{fig:framework}(a.1)). Each scenario undergoes $50$ independent calibrations to assess both accuracy and convergence. In each trial, the GT DH parameters are sampled from a uniform distribution, as listed in Table \ref{table:sim_param}. This random extrinsic configuration covers a range of realistic mounting conditions. The initial estimates are obtained by adding zero-mean Gaussian noise to the GT values. Such noise levels (up to $15^\circ$ in rotation and $0.15m$ in translation) emulate potential installation errors and mechanical deformation in real-world scenarios. Fig.~\ref{fig:monte_carlo_calib}  summarizes the error distribution across the simulations, where the translation error is below $1.5mm$, and the angle error is less than $0.04^\circ$. These results demonstrate that LM-Calibr achieves accuracy and convergence across various scenarios, initial parameters, and mounting configurations.

\begin{figure}[t]
\centering
\includegraphics[width=0.45\textwidth]{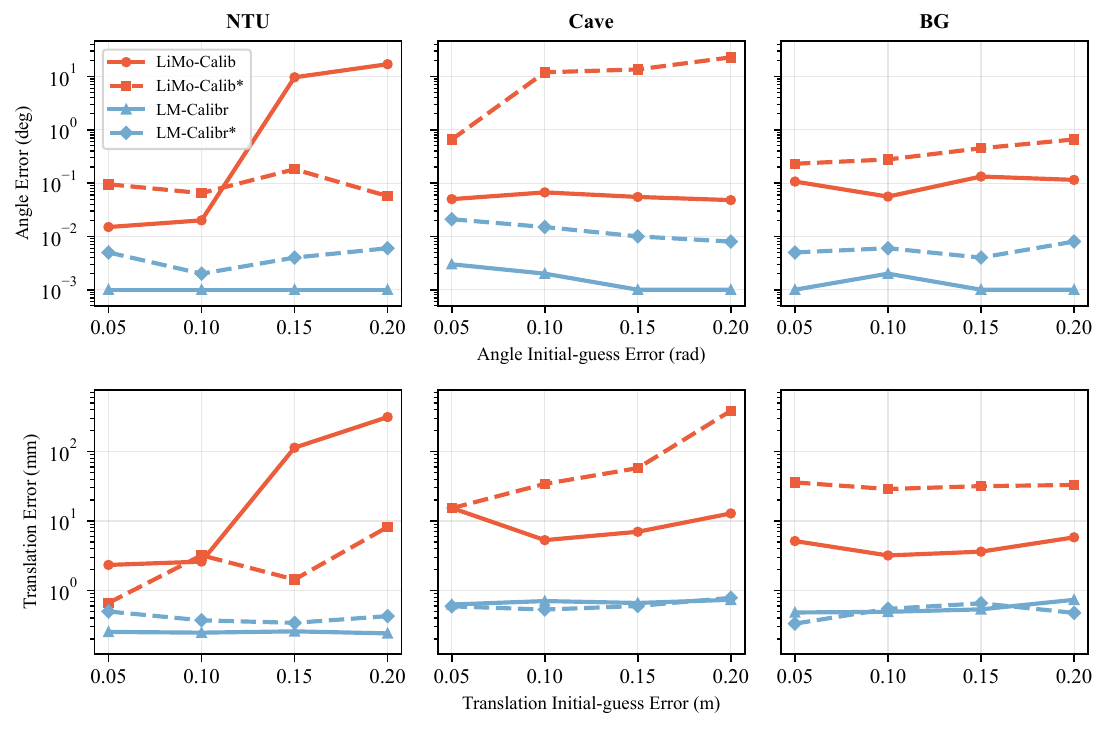}
\caption{The calibration error under different sensor types, scenarios, and initial-guess errors, where “*” denotes the method evaluated in Avia.}
\label{fig:limo_lm_compare}
\end{figure}

\subsubsection{Evaluate Accuracy in Simulation}

This subsection evaluates the accuracy of LM-Calibr and LiMo-Calib with synthetic data under varying noise levels. The GT extrinsic parameters are configured at positions where both methods can function properly. Different levels of noise are then added to the GT rotation and translation components to generate initial estimates. Fig.~\ref{fig:limo_lm_compare} presents the angle and translation errors under different sensor types, scenarios, and initial-guess errors. As the initial-guess errors magnitude increases, LM-Calibr maintains stable performance with angle error below $0.04^\circ$ and translation error less than $1 mm$. Conversely, the calibration error of LiMo-Calib deteriorates with elevated noise levels and exhibits significant performance degradation under high-error conditions ($0.2 rad$ in angle and $0.2 m$ in translation). These results manifest that LM-Calibr offers superior accuracy and enhanced convergence stability across all tested conditions.

\begin{figure}[t]
\centering
\includegraphics[width=0.45\textwidth]{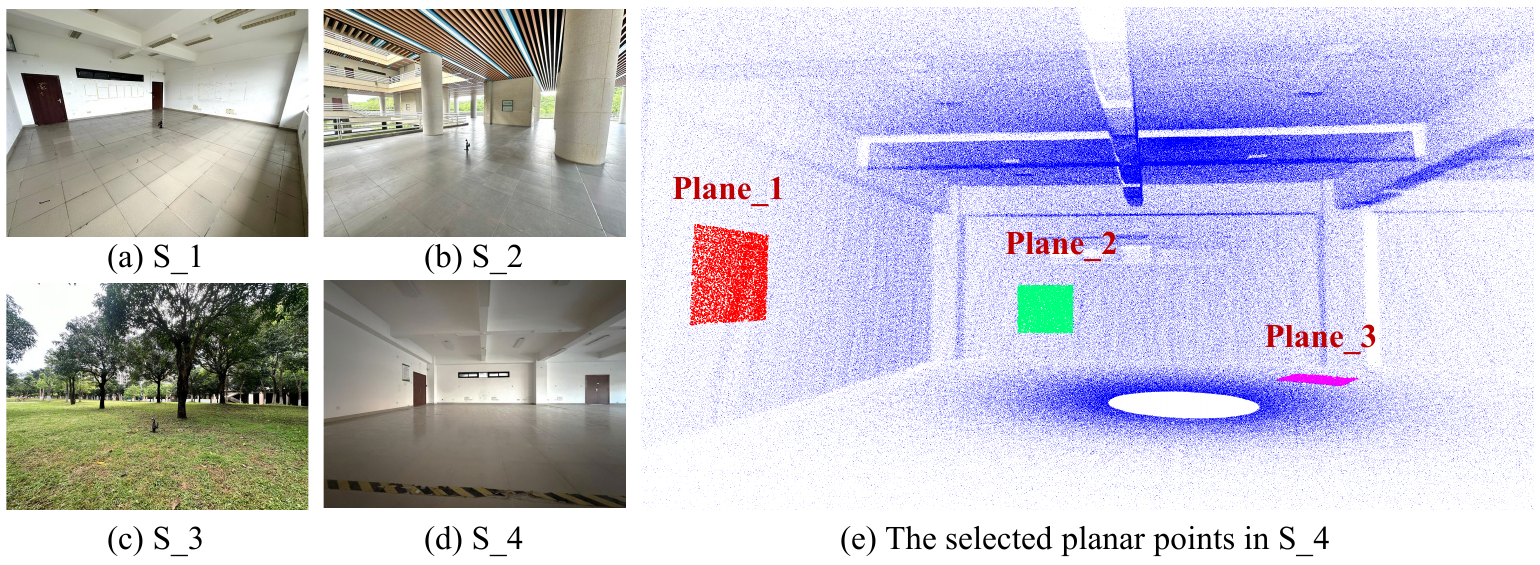}
\caption{(a)-(d) show the geometric structure of the real-world scenes. (e) illustrate the selected planar points for evaluating the point-to-plane distances.}
\label{fig:real_calib_plane}
\end{figure}

\subsubsection{Evaluate Accuracy in Real-world}
\begin{figure}[t]
\centering
\includegraphics[width=0.45\textwidth]{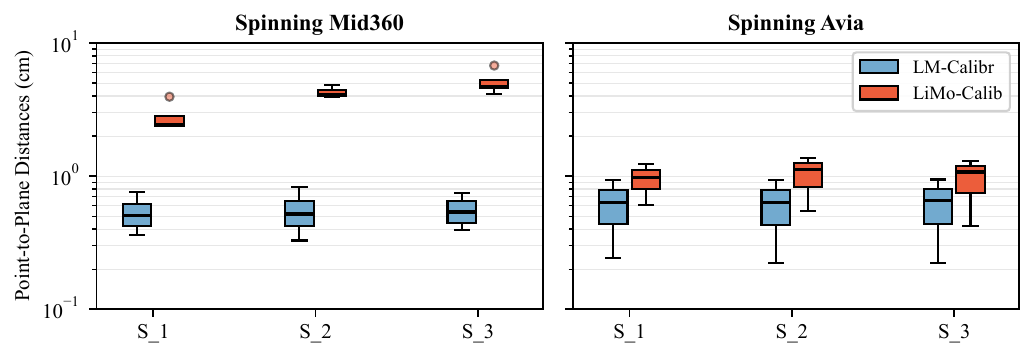}
\caption{The distributions of point-to-plane distances for LM-Calibr and LiMo-Calib in \texttt{S\_4}.}
\label{fig:plane_error_comparison}
\end{figure}

This subsection validates LM-Calibr and LiMo-Calib in three real-world scenes: structured (\texttt{S\_1}), semi-structured (\texttt{S\_2}), and forested (\texttt{S\_3}), as shown in Fig.~\ref{fig:real_calib_plane}(a-c). For each scene, four sequences are collected from different viewpoints. Each method is executed five times per sequence, yielding 20 calibration trials per scene.  The calibration results are subsequently evaluated via point-to-plane distances across three orthogonal planes ($1 m \times 1 m$) in an additional structured scene (\texttt{S\_4}), as illustrated in Fig.~\ref{fig:real_calib_plane}(d-e). The distances of LM-Calibr are less than $1 cm$, which are comparable to the LiDAR's intrinsic measurement errors, as presented in Fig.~\ref{fig:plane_error_comparison}. Conversely, LiMo-Calib suffers from significant errors and degeneracy because Mid360's x-axis aligns with the spinning axis. Moreover,  LM-Calibr outperforms LiMo-Calib by $20\%$ in accuracy, even under non-degenerate conditions for the spinning Avia calibration. These results demonstrate that LM-Calibr maintains robustness across diverse scenes and sensor configurations. 

\subsubsection{Efficiency}

\begin{figure}[t]
\centering
\includegraphics[width=0.45\textwidth]{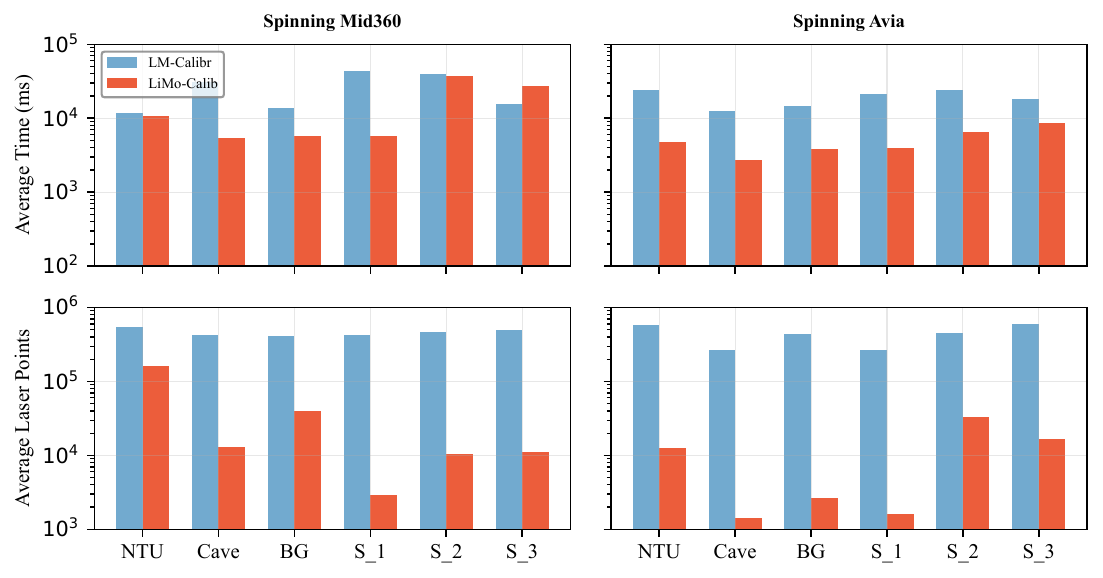}
\caption{The comparison between LM-Calibr and LiMo-Calib in average time and average laser points across different sequences.}
\label{fig:time_point_num}
\end{figure}

Fig.~\ref{fig:time_point_num} summarizes the LM-Calibr and LiMo-Calib efficiency across different sequences. LiMo-Calib improves efficiency by utilizing fewer laser points, but at the cost of reduced accuracy. The LM optimization in LM-Calibr is the most time-consuming step, as it requires computing point-wise Jacobians for all plane features. Therefore, in the structured (\texttt{S\_1}) and semi-structured (\texttt{S\_2}) environments, many laser points correspond to the same plane feature,  increasing the computational cost.

\begin{table*}[tb]
\caption{Absolute Trajectory Error (ATE, m), End-to-end Error (E2E, m), Time Consumption (ms) and Memory Usage (Gb) of LIOs}\label{table:ape_time_mem}
\centering
\tiny
\setlength{\tabcolsep}{1.7mm}
\begin{threeparttable}
\begin{tabular}{c|c|ccc|ccc|ccc|ccc|ccc|ccc|ccc}
\toprule[1pt]
                             &                        & \multicolumn{3}{c|}{EVA-LIO}                                                                                        & \multicolumn{3}{c|}{Voxel-SLAM}                                                             & \multicolumn{3}{c|}{Point-LIO}                                                                  & \multicolumn{3}{c|}{Fast-LIO2}                                                                  & \multicolumn{3}{c|}{Ada-LIO}                                                   & \multicolumn{3}{c|}{CTE-MLO}                                                   & \multicolumn{3}{c}{Traj-LO}                                                   \\
\multirow{-2}{*}{Eva. Env.}  & \multirow{-2}{*}{Seq.} & ATE                                             & Time                                  & Mem.                      & ATE                     & Time                                  & Mem.                      & ATE                         & Time                                  & Mem.                      & ATE                         & Time                                  & Mem.                      & ATE                     & Time                     & Mem.                      & ATE                     & Time                     & Mem.                      & ATE                     & Time                     & Mem.                     \\ \hline
                             & mid\_CA                & \textbf{0.11}                                   & 20.48                                 & 9.19                      & 0.20                    & {\color[HTML]{FE0000} \textbf{15.31}} & 4.68                      & 0.27                        & {\color[HTML]{0000FF} \textbf{19.2}}  & 0.38                      & 0.12                        & 23.49                                 & 0.54                      & 0.13                    & 49.84                    & 1.87                      & -\tnote{1}                       & -                        & -                         & 0.24                    & 48.32                    & \textbf{0.28}            \\
                             & mid\_CR                & \textbf{0.13}                                   & {\color[HTML]{0000FF} \textbf{14.69}} & 0.74                      & 0.19                    & {\color[HTML]{FE0000} \textbf{11.41}} & 1.52                      & 0.26                        & 42.71                                 & \textbf{0.32}             & 0.20                        & 16.37                                 & 0.44                      & 0.23                    & 57.53                    & 0.35                      & -                       & -                        & -                         & -                       & -                        & -                        \\
                             & mid\_BG                & \textbf{0.27}                                   & {\color[HTML]{0000FF} \textbf{26.19}} & 5.57                      & 0.27                    & {\color[HTML]{FE0000} \textbf{18.49}} & 4.79                      & 0.75                        & 34.97                                 & \textbf{0.53}             & 1.45                        & 31.09                                 & 0.60                      & 0.46                    & 52.98                    & 2.45                      & -                       & -                        & -                         & 0.71                    & 58.14                    & 0.98                     \\
                             & mid\_NTU               & \textbf{0.20}                                   & {\color[HTML]{0000FF} \textbf{32.96}} & 9.19                      & 0.21                    & 46.04                                 & 11.13                     & 0.97                        & {\color[HTML]{FE0000} \textbf{32.66}} & \textbf{0.80}             & 0.48                        & 36.17                                 & 1.10                      & 0.57                    & 59.09                    & 6.70                      & 0.41                    & 53.15                    & 1.00                      & 1.39                    & 54.55                    & 1.05                     \\
                             & mid\_KTH               & \textbf{0.16}                                   & {\color[HTML]{0000FF} \textbf{29.46}} & 8.05                      & 0.17                    & 29.81                                 & 8.97                      & 1.01                        & 34.60                                 & 0.87                      & 0.18                        & {\color[HTML]{FE0000} \textbf{24.01}} & 0.81                      & 9.02                    & 56.35                    & 6.81                      & 0.2                     & 42.77                    & \textbf{0.47}             & 0.74                    & 56.81                    & 1.39                     \\
\multirow{-6}{*}{Sim.}       & mid\_TUHH              & \textbf{0.07}                                   & {\color[HTML]{0000FF} \textbf{31.58}} & 7.85                      & 0.13                    & 38.25                                 & 9.29                      & 0.34                        & 35.24                                 & 0.81                      & 0.14                        & {\color[HTML]{FE0000} \textbf{26.00}} & 0.87                      & 0.14                    & 54.54                    & 6.86                      & 0.13                    & 45.54                    & \textbf{0.25}             & 0.58                    & 56.98                    & 1.43                     \\ \hline
                             & avia\_CA               & \textbf{0.25}                                   & {\color[HTML]{0000FF} \textbf{9.59}}  & 1.14                      & 2.55                    & {\color[HTML]{FE0000} \textbf{7.78}}  & 2.83                      & 3.31                        & 16.59                                 & \textbf{0.38}             & -                           & -                                     & -                         & -                       & -                        & -                         & -                       & -                        & -                         & -                       & -                        & -                        \\
                             & avia\_CR               & \textbf{0.36}                                   & {\color[HTML]{0000FF} \textbf{7.67}}  & \textbf{0.75}             & 0.36                    & {\color[HTML]{FE0000} \textbf{6.46}}  & 0.99                      & -                           & -                                     & -                         & -                           & -                                     & -                         & -                       & -                        & -                         & -                       & -                        & -                         & -                       & -                        & -                        \\
                             & avia\_BG               & \textbf{0.15}                                   & {\color[HTML]{0000FF} \textbf{13.77}} & 4.63                      & 0.39                    & {\color[HTML]{FE0000} \textbf{9.29}}  & 3.08                      & 2.75                        & 18.21                                 & \textbf{0.48}             & -                           & -                                     & -                         & -                       & -                        & -                         & -                       & -                        & -                         & -                       & -                        & -                        \\
                             & avia\_NTU              & \textbf{0.54}                                   & {\color[HTML]{0000FF} \textbf{29.49}} & 8.96                      & 0.59                    & 29.55                                 & 8.02                      & 1.08                        & 34.4                                  & \textbf{0.61}             & 8.41                        & {\color[HTML]{FE0000} \textbf{26.09}} & 0.65                      & -                       & -                        & -                         & -                       & -                        & -                         & -                       & -                        & -                        \\
                             & avia\_KTH              & \textbf{0.37}                                   & 18.74                                 & 7.42                      & 0.43                    & {\color[HTML]{FE0000} \textbf{17.88}} & 4.80                      & 1.14                        & 24.01                                 & 0.79                      & 2.24                        & {\color[HTML]{0000FF} \textbf{18.23}} & \textbf{0.74}             & -                       & -                        & -                         & -                       & -                        & -                         & -                       & -                        & -                        \\
\multirow{-6}{*}{Sim.}       & avia\_THUU             & \textbf{0.42}                                   & {\color[HTML]{0000FF} \textbf{21.83}} & 7.46                      & 0.44                    & 23.97                                 & 6.51                      & 0.60                        & 38.71                                 & \textbf{0.80}             & 0.86                        & {\color[HTML]{FE0000} \textbf{18.69}} & 0.84                      & -                       & -                        & -                         & -                       & -                        & -                         & -                       & -                        & -                        \\ \hline
                             & Seq.                   & \multicolumn{1}{l}{E2E}                         & \multicolumn{1}{l}{Time}              & \multicolumn{1}{l|}{Mem.} & \multicolumn{1}{l}{E2E} & \multicolumn{1}{l}{Time}              & \multicolumn{1}{l|}{Mem.} & \multicolumn{1}{l}{E2E}     & \multicolumn{1}{l}{Time}              & \multicolumn{1}{l|}{Mem.} & \multicolumn{1}{l}{E2E}     & \multicolumn{1}{l}{Time}              & \multicolumn{1}{l|}{Mem.} & \multicolumn{1}{l}{E2E} & \multicolumn{1}{l}{Time} & \multicolumn{1}{l|}{Mem.} & \multicolumn{1}{l}{E2E} & \multicolumn{1}{l}{Time} & \multicolumn{1}{l|}{Mem.} & \multicolumn{1}{l}{E2E} & \multicolumn{1}{l}{Time} & \multicolumn{1}{l}{Mem.} \\ \cline{2-23} 
                             & mid\_real              & {\color[HTML]{000000} \textbf{\textless{}0.01}} & {\color[HTML]{FE0000} \textbf{27.07}} & 7.36                      & 3.54                    & {\color[HTML]{0000FF} \textbf{28.52}} & 8.56                      & -                           & -                                     & -                         & {\color[HTML]{000000} 2.80} & 32.10                                 & \textbf{0.85}             & 3.00                    & 54.89                    & 9.86                      & -                       & -                        & -                         & -                       & -                        & -                        \\
\multirow{-3}{*}{Real-World} & avia\_real             & {\color[HTML]{000000} \textbf{\textless{}0.01}} & {\color[HTML]{0000FF} \textbf{16.09}} & 2.06                      & 12.29                   & 29.19                                 & 4.44                      & {\color[HTML]{000000} 0.07} & 36.99                                 & 0.45                      & {\color[HTML]{333333} 0.02} & {\color[HTML]{FE0000} \textbf{15.49}} & \textbf{0.32}             & -                       & -                        & -                         & -                       & -                        & -                         & -                       & -                        & -                        \\ \bottomrule[1pt]
\end{tabular}
Bold black highlights the minimum ATE and memory usage. Bold red indicates the minimum time consumption, while bold blue denotes the second minimum. 
\tnote{1} “-” denotes that the system failed. The baseline code and config will be provided on GitHub.
\end{threeparttable}
\end{table*}

\begin{figure*}[t]
\centering
\includegraphics[width=0.95\textwidth]{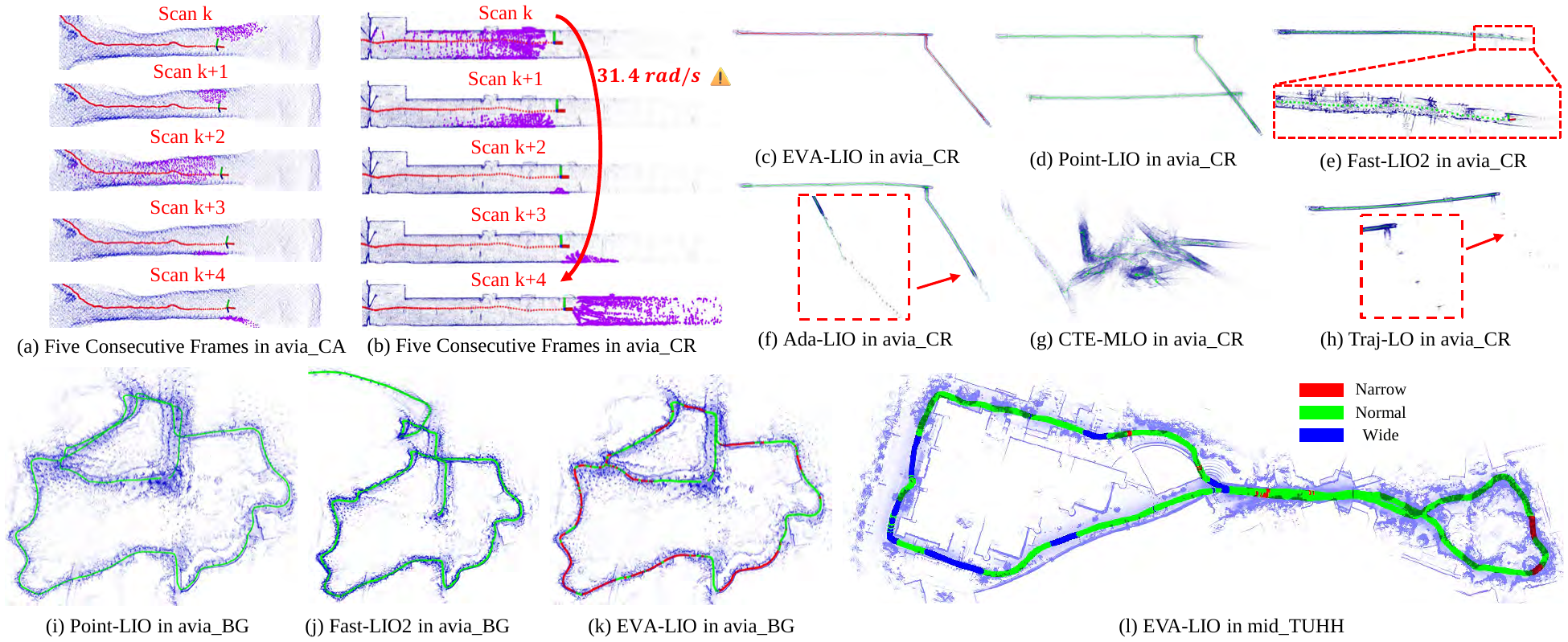}
\caption{The mapping results of different LIOs in the synthetic sequences. (a) and (b) illustrate five consecutive frames in \texttt{avia\_CA} and \texttt{avia\_CR}. The purple laser points indicate Avia's scan pattern. (c)-(h) present the mapping results of different methods in \texttt{avia\_CR}. Notably, Point-LIO, Fast-LIO2, Ada-LIO, CTE-MLO, and Traj-LO exhibit significant drift in their generated maps. (i) and (j) show severe misalignment mapping results of Point-LIO and Fast-LIO2 in  \texttt{avia\_BG}. (k) and (l) describe the global maps of EVA-LIO in \texttt{avia\_BG} and \texttt{mid\_TUHH}. The trajectories, represented by different colors, correspond to various spatial scales identified by the environmental analysis module. Red, green, and blue denote narrow, normal, and wide types, respectively.}
\label{fig:sim_loc_results}
\end{figure*}

\subsection{The Accuracy and Robustness of EVA-LIO}\label{sec:eva_lio_exp}

To assess the accuracy and robustness, EVA-LIO is compared with state-of-the-art (SOTA) LIO systems:  Voxel-SLAM \cite{voxel_slam}, Point-LIO \cite{point_lio}, Fast-LIO2 \cite{fast_lio_2}, Adaptive-LIO (Ada-LIO) \cite{adaptive_lio}, CTE-MLO \cite{cte_mlo}, and Traj-LO \cite{traj_lo}. Point-LIO employs point-wise updates for accurate pose estimation under aggressive motion and has been deployed on the Unitree-L2 spinning LiDAR\footnote{https://github.com/unitreerobotics/point\_lio\_unilidar}. Ada-LIO is a related work that manages multi-resolution voxel maps to improve localization accuracy and mitigate degeneration. CTE-MLO and Traj-LO are SOTA continuous-time LiDAR odometry. Voxel-SLAM is a complete and versatile LiDAR-inertial SLAM system with an adaptive voxel map. For fair comparison, only the LIO component of Voxel-SLAM is enabled. Additionally, $\alpha$LiDAR and UA-MPC are omitted, as their motor controllers are not open-source.

EVA-LIO utilizes consistent parameters across all sequences. Specifically, the root sizes of the voxel maps $\mathcal{M}_1, \mathcal{M}_2$, and $\mathcal{M}_3$ are set to be $0.25m, 0.5m$, and $1.0m$, respectively. Each voxel map has a maximum of two layers. The downsample rates $V_1, V_2$, and $V_3$ are $0.15m, 0.2m$, and $0.25m$, respectively. The system parameters of baseline LIOs are listed in Section G of the supplementary material. All the compared methods apply the IMU in the base frame as input.

\subsubsection{Evaluation in Simulation}

The experiments utilize \verb|Cave04| (CA) \cite{super_loc}, \verb|Corridor01| (CR) \cite{super_loc}, \verb|botanic_1006_01 | (BG) \cite{botanic_garden}, \verb|NTU_01| (NTU) \cite{mcd}, \verb|KTH_06| (KTH) \cite{mcd}, and \verb|TUHH_03| (TUHH) \cite{mcd} as MARSIM inputs to generate spinning Mid360 and spinning Avia datasets. The motor speed is set to be $7.85 rad/s$ for all sequences. The absolute trajectory errors are reported in Table \ref{table:ape_time_mem}. 

CA is a challenging cave environment with narrow passages and repetitive structures. When the spinning Avia approaches the rock wall, LIO fails to extract sufficient geometric features, as illustrated in Fig.~\ref{fig:sim_loc_results}(a). As a result, Voxel-SLAM, Point-LIO, Fast-LIO2, Ada-LIO, CTE-MLO, and Traj-LO show degraded performance in \texttt{avia\_CA}. In contrast, EVA-LIO remains robust, even when LiDAR briefly scans featureless areas.

CR is a long corridor with repetitive and symmetrical structures. Fig.~\ref{fig:sim_loc_results}(b) shows Avia's scans from five consecutive frames in \texttt{avia\_CR}, where Avia's scans temporarily cover the featureless areas. For the UA-MPC strategy \cite{ua_mpc}, it would require a motor speed of $31.4 rad/s$, which is beyond the dynamic feasibility of the motor. Due to a lack of distinct geometric features, Point-LIO, Fast-LIO2, Ada-LIO, CTE-MLO, and Traj-LO show divergence and drift in \texttt{avia\_CR}, as present in Fig.~\ref{fig:sim_loc_results}(d)-(h). EVA-LIO and Voxel-SLAM estimate per-point uncertainty, which improves robustness in such extreme scenarios, as illustrated in Fig.~\ref{fig:sim_loc_results}(c) and Table \ref{table:ape_time_mem}.

BG is collected in a dense botanical garden with repetitive patterns and unstructured environments. Due to degraded feature matching, Ada-LIO, CTE-MLO and Traj-LO exhibit significant drift or divergence in BG. Point-LIO and Fast-LIO2 produce misaligned maps in \texttt{avia\_BG}, as shown in Fig.~\ref{fig:sim_loc_results}(i) and (j). In comparison, EVA-LIO demonstrates accuracy and robustness in this scenario. This exceptional performance is attributed to per-point uncertainty estimation and the environmental analysis module, as described in Fig.~\ref{fig:sim_loc_results}(k).

NTU, KTH, and TUHH are collected from three Eurasian university campuses with diverse environmental scales and geometric feature distributions. EVA-LIO can adaptively select the downsample rate $V_{s}$ and voxel map $\mathcal{M}_{s}$ based on the spatial scale, resulting in higher accuracy than Voxel-SLAM, as illustrated in Fig.~\ref{fig:sim_loc_results}(l) and Table \ref{table:ape_time_mem}. All the simulation results confirm that EVA-LIO achieves accuracy and robustness across all sequences, regardless of the different LiDAR types.

\begin{figure}[t]
\centering
\includegraphics[width=0.45\textwidth]{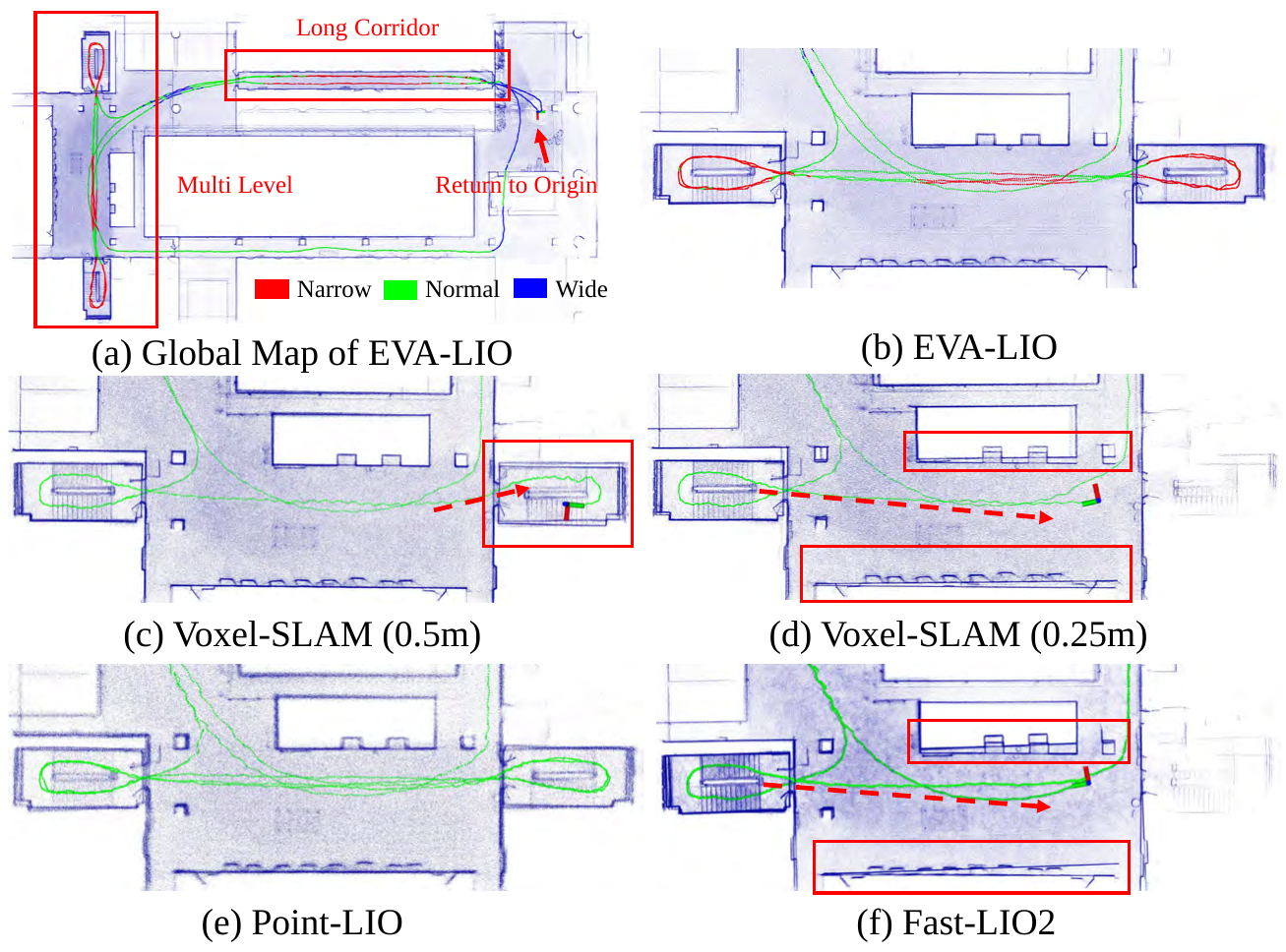}
\caption{The mapping results of different methods in \texttt{avia\_real}. (a) presents the global trajectory of EVA-LIO, with red, green, and blue colors indicating different environment types. (b) illustrates the map of EVA-LIO at the junction of the stairwell and hallway. (c) and (d) depict the maps of Voxel-SLAM with the root size of $0.5m$ and $0.25m$. Dashed arrows indicate the device's movement direction. Voxel-SLAM (0.5) drifts in the stairwell, while Voxel-SLAM (0.25) drifts in the hallway. (e) shows the map of Point-LIO with visible ghosting. (f) highlights significant distortion of Fast-LIO2 in the hallway.}
\label{fig:real_loc_avia}
\end{figure}

\begin{figure}[t]
\centering
\includegraphics[width=0.45\textwidth]{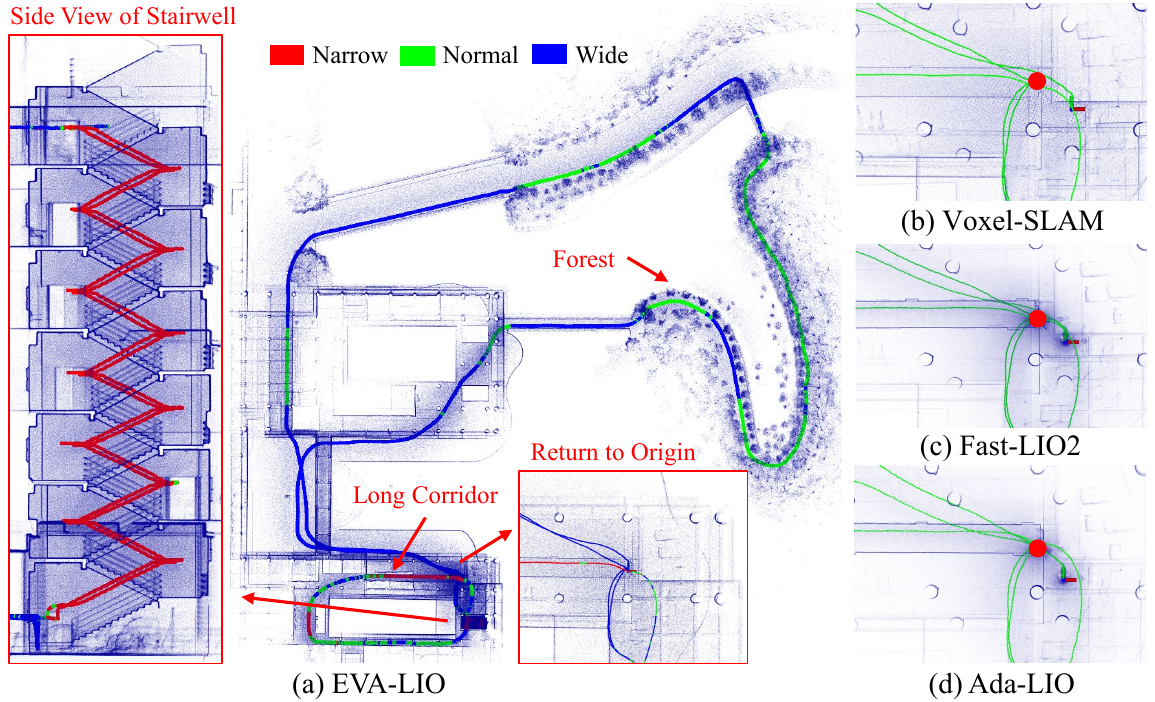}
\caption{The mapping results of different methods in \texttt{mid\_real}. (a) shows the global trajectory and local mapping details of EVA-LIO. The red, green, and blue points in the trajectory represent different environment types. (b), (c) and (d) illustrate that Voxel-SLAM,  Fast-LIO2, and Ada-LIO fail to return to the origin point (red point) after long-term localization, respectively.}
\label{fig:real_loc_mid}
\end{figure}

\subsubsection{Evaluation in Real-world}

For the real-world experiment, \texttt{avia\_real} and \texttt{mid\_real} are collected with the custom sensing system in Fig.~\ref{fig:framework}(a). All sensors are hardware-synchronized, and the system design is open-sourced on GitHub. 
Both sequences follow closed-loop trajectories where the start and end points coincide. Table \ref{table:ape_time_mem} reports the end-to-end errors for the evaluated LIOs. In \texttt{avia\_real}, the motor operates at a maximum speed of $7.85 rad/s$. This sequence includes challenge scenes such as a long corridor and stairwells, as depicted in Fig~\ref{fig:real_loc_avia}(a). In the stairwells, the LiDAR's scan temporarily covers areas with few features. Additionally, transitioning between narrow and wide scenes, such as exiting or entering the stairwells, poses challenges to localization robustness. As a result, Ada-LIO, CTE-MLO, and Traj-LO diverge in the long corridor. The maps built by Voxel-SLAM, Point-LIO, and Fast-LIO2 exhibit inconsistencies. Although the adaptive voxel map in Voxel-SLAM can subdivide voxels based on local geometric features, it still suffers from drift in the stairwell and hallway when the root sizes are set to be $0.5 m$ and $0.25 m$, respectively. In contrast, the environmental analysis module selects the narrow map $\mathcal{M}_{1}$ and the normal map $\mathcal{M}_{2}$ for localization in the stairwell and hallway, ensuring accurate mapping, as shown in Fig.~\ref{fig:real_loc_avia}(b).

In \texttt{mid\_real}, the motor operates at a maximum speed of $14 rad/s$. The operator traverses a seven-story building with the handheld device, passes through a forest, and then returns to the origin. The total trajectory length is $1104 m$ and contains various elements, including multi-level structures, a long corridor, semi-structured, and unstructured environments. Fig.~\ref{fig:real_loc_mid}(a) presents the global trajectory and mapping details of EVA-LIO. Conversely, Point-LIO and CTE-MLO degenerate in the long corridor, and Traj-LO diverge in the stairwell. Voxel-SLAM, Fast-LIO2, and Ada-LIO fail to return to the origin after long-term mapping, as shown in Fig.~\ref{fig:real_loc_mid}(b), (c), and (d), respectively.

Both the simulation and real-world experiments demonstrate that the environmental analysis module is able to select an appropriate downsample rate and voxel map based on the spatial scale. This adaptability enables EVA-LIO to achieve accuracy and robust localization across various scenarios, even when the LiDAR briefly scans featureless areas.

\subsubsection{Efficiency and Memory Usage}

Table \ref{table:ape_time_mem} summarizes the runtime and memory usage of each method across different sequences. Despite performing environmental analysis and managing multiple voxel maps, EVO-LIO consistently remains real-time performance on embedded platforms. In certain sequences, EVA-LIO consumes less memory than Voxel-SLAM, primarily due to its optimized map data structure, which reduces redundant storage.

\subsection{Ablation Study}

This subsection validates the contributions of uncertainty estimation and spinning actuation by evaluating EVA-LIO (w/o spin.) and EVA-LIO (w/o uncert.) on CR sequences. As detailed in Table \ref{table:ablation_study}, the ablation of these components leads to a marked decline in accuracy compared to the full system results in Table \ref{table:ape_time_mem}. Specifically, the Avia frequently scans featureless regions in CR sequences. EVA-LIO inevitably suffers from estimation degeneration in the absence of uncertainty propagation or spinning actuation to expand the FOV. Furthermore, evaluations on the MCD benchmark (see Section E of the supplementary material) demonstrate that the proposed uncertainty estimation and environmental analysis modules enable EVA-LIO (w/o spin.) to outperform baseline LIOs in localization accuracy.

\section{Failure Case and Limitations}

Although LM-Calibr is a targetless calibration method, its underlying principle is to employ adaptive voxelization to partition the environment and extract local planes. Consequently, its performance depends on both the spatial distribution and the number of LiDAR points. Notably, LM-Calibr ensures the identifiability even in extreme cases, e.g., when only ground points are available (see Section D of the supplementary material). For EVA-LIO, the system exhibits robustness against high motor speeds and rapid motor acceleration changes. However, its performance degrades in highly dynamic scenarios, e.g., maneuvers with accelerations $\ge 6g$ (see Section F of the supplementary material).

\begin{table}[!t]
\caption{Absolute Trajectory Error (m) of Ablation Study}\label{table:ablation_study}
\setlength{\tabcolsep}{4.3mm}
\centering
\scriptsize
\begin{threeparttable}
\begin{tabular}{ccc}
\toprule[1pt]
Seq.     & EVA-LIO (w/o uncert.) & EVA-LIO (w/o spin.) \\ \hline
mid\_CR  & 0.17                  & 0.32                \\
avia\_CR & -\tnote{1}                     & - \\ \bottomrule[1pt]                  
\end{tabular}
\tnote{1} “-” denotes that the system failed.
\end{threeparttable}
\end{table}

\section{Conclusions}\label{sec:conclusions}
This letter proposed LM-Calibr and EVA-LIO for accurate calibration and robust localization of spinning actuated LiDAR systems. LM-Calibr parametrized the extrinsic parameters of the LiDAR-motor systems via the DH convention and calibrated the DH parameters through the coarse-to-fine optimization. Extensive experiments showed that LM-Calibr did not rely on special target objects and achieved accurate calibration across different mountings and initial guesses. EVA-LIO modeled uncertainty for each laser point and managed multiple adaptive voxel maps. The environmental analysis module enabled EVA-LIO to select the appropriate downsample rate and voxel map. Experiments demonstrated that EVA-LIO maintained robust localization in various challenging scenarios, even when the actuator operated at maximum speed.





\section{Supplementary Material}

Please note that equation numbers and section numbers from the main manuscript are labelled in this letter in \textcolor{red}{red}.

\subsection{Denavit-Hartenberg Parameters Calibration} \label{sec:supp_dh_calib}

The cost function of LM-Calibr is defined as
\begin{equation}\label{eq:supp_lm_calibr_cost_func}
\min_{\mathbf{x}}(\sum^{M_f}_{i=1}\pi_{i}(\mathbf{x})),
\end{equation}
where
\begin{equation*}
\begin{split}
\mathbf{x} &\triangleq  \begin{bmatrix}
\bar{\theta} & \bar{d} & \bar{a} & \bar{\phi}
\end{bmatrix}^\top, \\
\pi_i(\mathbf{x}) &= \lambda_{\min}(\mathbf{A}_i(\mathbf{x})), \\
\mathbf{A}_i(\mathbf{x}) &= \frac{1}{N_i}\sum^{N_i}_{j=1}(\mathbf{p}^M_{ij}(\mathbf{x}) - \mathbf{q}_i(\mathbf{x}) ) ( \mathbf{p}^M_{ij}(\mathbf{x}) - \mathbf{q}_i(\mathbf{x}) )^\top, \\
\mathbf{q}_i(\mathbf{x}) &= \frac{1}{N_i}\sum^{N_i}_{j=1}\mathbf{p}^M_{ij}(\mathbf{x}), \mathbf{p}^M_{ij}(\mathbf{x}) = \mathbf{T}^M_L(\mathbf{x}) \mathbf{p}^L_{ij}.\\ 
\end{split}
\end{equation*}
The $i$-th plane feature $\pi_i$ contains $N_i$ laser points. $\mathbf{p}^L_{ij}$ represents the $j$-th laser point in the $i$-th plane feature. $\lambda_{\min}(\mathbf{A})$ denotes the minimum eigenvalue of matrix $\mathbf{A}$. The Jacobian matrix $\mathbf{J}$ and the Hessian matrix $\mathbf{H}$ of the cost function \eqref{eq:supp_lm_calibr_cost_func} are given by
\begin{equation}
\begin{split}
\mathbf{J}_{ij} &= \frac{\partial\lambda_{\text{min}}(\mathbf{A}_i(\mathbf{x}))}{\partial \mathbf{p}^M_{ij}(\mathbf{x})}\frac{\partial \mathbf{p}^M_{ij}(\mathbf{x})}{\partial \delta \mathbf{x}}, \\
\mathbf{H}_{ijk} &= \bigg(\frac{\partial \mathbf{p}^M_{ij}(\mathbf{x})}{\partial \delta \mathbf{x}}\bigg)^\top \frac{\partial^2\lambda_{\min}(\mathbf{A}_i(\mathbf{x}))}{\partial \mathbf{p}^M_{ij}(\mathbf{x}) \partial \mathbf{p}^M_{ik}(\mathbf{x})} \frac{\partial \mathbf{p}^M_{ik}(\mathbf{x})}{\partial \delta \mathbf{x}}, \\
\end{split}
\end{equation}
where the first and second order derivatives of $\lambda_{\text{min}}(\mathbf{A}_i(\mathbf{x}))$ are provided in \cite{balm_1}. The derivative of the laser point is expressed as
\begin{equation}
\frac{\partial \mathbf{p}^M_{ij}(\mathbf{x})}{\partial \delta \mathbf{x}} = \begin{bmatrix}
\frac{\partial \mathbf{p}^M_{ij}(\mathbf{x})}{\partial \delta \bar{\theta}} & \frac{\partial \mathbf{p}^M_{ij}(\mathbf{x})}{\partial \delta \bar{d}} & \frac{\partial \mathbf{p}^M_{ij}(\mathbf{x})}{\partial \delta\bar{a}} & \frac{\partial \mathbf{p}^M_{ij}(\mathbf{x})}{\partial \delta \bar{\phi}}
\end{bmatrix}.
\end{equation}
The first two components are obtained as
\begin{align}
\frac{\partial  \mathbf{p}^M_{ij}(\mathbf{x})}{\partial \delta \bar{\theta}}=& \frac{\partial  \mathbf{p}^M_{ij}(\mathbf{x})}{\partial \delta \theta_2} \\
=&\bigg[ -\mathbf{R}_z(\theta_j)\mathbf{R}_x(\phi_1)\mathbf{R}_z(\theta_2)[\mathbf{R}_x(\phi_2)\mathbf{p}^L_{ij} + \mathbf{t}_1]_{\times} \bigg ]_3, \nonumber\\
\frac{\partial \mathbf{p}^M_{ij}(\mathbf{x})}{\partial \delta \bar{d}} =& \frac{\partial \mathbf{p}^M_{ij}(\mathbf{x})}{\partial \delta d_2}= \bigg[ \mathbf{R}_z(\theta_j) \mathbf{R}_x(\phi_1) \mathbf{R}_z(\theta_2)\bigg]_{3}, \nonumber
\end{align}
where $[\cdot]_{\times}$ is the skew-symmetric operator. $[\cdot]_{n}$ stands for the $n$-th column of a matrix. Angle $\theta_j$ denotes the interpolated encoder measurement of the $j$-th laser point. The final two terms depend on the LiDAR type. For the spinning omni LiDAR, the derivatives are
\begin{align}
\frac{\partial  \mathbf{p}^M_{ij}(\mathbf{x})}{\partial \delta \bar{a}} &= \frac{\partial  \mathbf{p}^M_{ij}(\mathbf{x})}{\partial \delta a_1}= \bigg[ \mathbf{R}_z(\theta_j) \bigg]_1, \\
\frac{\partial  \mathbf{p}^M_{ij}(\mathbf{x})}{\partial \delta \bar{\phi}} &= \frac{\partial  \mathbf{p}^M_{ij}(\mathbf{x})}{\partial \delta \phi_1} \nonumber\\
&= \bigg[ -\mathbf{R}_z(\theta_j)\mathbf{R}_x(\phi_1)[\mathbf{R}_z(\theta_2)(\mathbf{R}_x(\phi_2)\mathbf{p}^L_{ij} + \mathbf{t}_1)]_{\times} \bigg]_1. \nonumber
\end{align}
For the spinning non-omni LiDAR, they are given by
\begin{align}
\frac{\partial \mathbf{p}^M_{ij}(\mathbf{x})}{\partial \delta \bar{a}} &= \frac{\partial \mathbf{p}^M_{ij}(\mathbf{x})}{\partial \delta a_2}= \bigg[ \mathbf{R}_z(\theta_j)\mathbf{R}_x(\phi_1)\mathbf{R}_z(\theta_2) \bigg]_1,  \\
\frac{\partial \mathbf{p}^M_{ij}(\mathbf{x})}{\partial \delta \bar{\phi}} &= \frac{\partial \mathbf{p}^M_{ij}(\mathbf{x})}{\partial \delta \phi_2} \nonumber\\
&= \bigg[ -\mathbf{R}_z(\theta_j)\mathbf{R}_x(\phi_1)\mathbf{R}_z(\theta_2)\mathbf{R}_x(\phi_2)[\mathbf{p}^L_{ij}]_{\times} \bigg]_1. \nonumber
\end{align}

\subsection{Probabilistic Laser Point Representation}\label{sec:supp_prob_laser_point}

This subsection derives a probabilistic model for the noise $\delta \mathbf{p}^W_j \sim \mathcal{N}(\mathbf{0}, \boldsymbol{\Sigma}_{\mathbf{p}^W_j})$ of a laser point $\mathbf{p}^W_j $, which incorporates measurement noise from both the LiDAR and the encoder.  Following the measurement noise model presented in \cite{pixel_lidar_cam_calib}, the noise $\delta \mathbf{p}_j^L$ associated with a laser point $\mathbf{p}^L_j$ is formulated as
\begin{align}
\delta \mathbf{p}^L_j &= \mathbf{A}_j \begin{bmatrix}
\delta d_j \\
\delta \boldsymbol{\omega}_j
\end{bmatrix} \sim \mathcal{N}(\mathbf{0}, \boldsymbol{\Sigma}_{\mathbf{p}^L_j}), \\
\boldsymbol{\Sigma}_{\mathbf{p}^L_j} &= \mathbf{A}_j
\begin{bmatrix}
\Sigma_{d_i} & \mathbf{0} \\
\mathbf{0} & \boldsymbol{\Sigma}_{\boldsymbol{\omega}_j}
\end{bmatrix}
\mathbf{A}_j^\top, \nonumber\\
\mathbf{A}_j &= \begin{bmatrix}
\boldsymbol{\omega}_j & -d_j[\boldsymbol{\omega}_j]_{\times} \mathbf{N}(\boldsymbol{\omega}_j)
\end{bmatrix},\nonumber
\end{align}
where the vector $\boldsymbol{\omega}_j\in \mathbb{S}^2$ denotes the bearing direction of the laser point, and $\delta\boldsymbol{\omega}_j \sim \mathcal{N}(\mathbf{0}, \boldsymbol{\Sigma}_{\boldsymbol{\omega}_j})$  represents the corresponding bearing direction noise. The scalar $d_j$ denotes the measured depth, with associated noise $\delta d_j \sim \mathcal{N}(0, \Sigma_{d_j})$. The matrix $\mathbf{N}(\boldsymbol{\omega}_j) = \begin{bmatrix} \mathbf{N}_1 & \mathbf{N}_2\end{bmatrix}$ forms an orthonormal basis of the tangent plane at $\boldsymbol{\omega}_j$. 

The laser point $\mathbf{p}^L_j$ is then transformed from the LiDAR frame $\{L\}$ to the motor frame $\{M\}$ via \textcolor{red}{Eq. \eqref{eq:full_rigid_trans}}. The laser point in $\{M\}$ is expressed as
\begin{equation}
\begin{split}
\mathbf{p}_j^M &= \mathbf{R}_z(\theta_j) ( \mathbf{R}_x(\phi_1)\mathbf{R}_z(\theta_2)(\mathbf{R}_x(\phi_2)\mathbf{p}^L_j + \mathbf{t_1}) + \mathbf{t}_2 ) \\
&= \mathbf{R}_z(\theta_j)(\bar{\mathbf{R}} \mathbf{p}^L_j + \bar{\mathbf{t}}),
\end{split}
\end{equation}
where
\begin{equation*}
\begin{split}
\bar{\mathbf{R}} &= \mathbf{R}_x(\phi_1)\mathbf{R}_z(\theta_2)\mathbf{R}_x(\phi_2), \\
\bar{\mathbf{t}} &= \mathbf{R}_x(\phi_1)\mathbf{R}_z(\theta_2)\mathbf{t}_1 + \mathbf{t}_2.
\end{split}
\end{equation*}
The motor angle $\theta_j$ is obtained via interpolation as
\begin{equation}
\begin{split}
\theta_j &= (1-\lambda)\theta_a + \lambda\theta_b \\
\lambda &= (t_j - t_a) / (t_b - t_a), t_a \leq t_j \leq t_b,
\end{split}
\end{equation}
where $\theta_a$ and $\theta_b$ are the encoder measurements at time instances adjacent to the $j$-th laser point. As discussed in \textcolor{red}{Section \ref{sec:ada_lio}}, the significant disparity in sampling frequencies between the LiDAR and the encoder compromises the accuracy of $\theta_j$, and thus its uncertainty must be explicitly modeled. Since microsecond-level time synchronization is available between the LiDAR and the encoder,  the uncertainty in $\theta_j$ primarily stems from the nonlinear motor’s speed variations within the sampling intervals. By modeling these errors as Gaussian noise $\delta \theta \sim \mathcal{N}(0,\Sigma_{\theta})$, the ground-truth laser point $\mathbf{p}^{M,gt}_j$ can be expanded as
\begin{align}
\mathbf{p}^{M,gt}_j =& \mathbf{R}_z((1-\lambda)\theta_a + \lambda\theta_b + \delta\theta)(\bar{\mathbf{R}} (\mathbf{p}^L_j + \delta \mathbf{p}^L_j )  + \bar{\mathbf{t}}) \nonumber\\
\approx&  \mathbf{p}_j^M + \mathbf{J}_{\theta} \delta \theta + \mathbf{J}_{\mathbf{p}^L_j} \delta \mathbf{p}^L_j,
\end{align}
where
\begin{equation*}
\begin{split}
\mathbf{J}_\theta =&   \bigg[ -\mathbf{R}_z(\theta_j)[\bar{\mathbf{R}}\mathbf{p}^L_j + \bar{\mathbf{t}}]_{\times} \bigg]_3,\\
\mathbf{J}_{\mathbf{p}^L_j} = &\mathbf{R}_z(\theta_j) \bar{\mathbf{R}}.
\end{split}
\end{equation*}
The operator $[\cdot]_n$ stands for the $n$-th column of a matrix. Consequently, the noise $\delta \mathbf{p}^M_j$ of the laser point $\mathbf{p}^M_j$ and its covariance $\boldsymbol{\Sigma}_{\mathbf{p^M_j}}$ is derived as 
\begin{align}
\delta \mathbf{p}^M_j &=  \begin{bmatrix}
\mathbf{J}_{\theta} &\mathbf{J}_{\mathbf{p}^L_j}
\end{bmatrix} 
\begin{bmatrix}
\delta \theta \\
\delta \mathbf{p}^L_j
\end{bmatrix} \sim \mathcal{N}(0, \boldsymbol{\Sigma}_{\mathbf{p^M_j}}),\\
\boldsymbol{\Sigma}_{\mathbf{p^M_j}} &= \begin{bmatrix}
\mathbf{J}_{\theta} &\mathbf{J}_{\mathbf{p}^L_j}
\end{bmatrix} 
\begin{bmatrix}
\Sigma_\theta & \mathbf{0} \\
\mathbf{0} & \boldsymbol{\Sigma}_{\mathbf{p}^L_j}
\end{bmatrix}
\begin{bmatrix}
\mathbf{J}_{\theta} &\mathbf{J}_{\mathbf{p}^L_j}
\end{bmatrix}^\top.\nonumber
\end{align}

The laser point $\mathbf{p}^M_j$ is further projected into the world frame ${W}$ as 
\begin{equation}
\mathbf{p}^W_j = \mathbf{R}^W_B (\mathbf{R}^B_M \mathbf{p}^M_j + \mathbf{t}^B_M) + \mathbf{t}^W_B,
\end{equation}
where the rotation $\mathbf{R}^W_B$ and translation $\mathbf{t}^W_B$ correspond to the state estimation from the LIO system. The rigid-body transformation parameters $\mathbf{R}^B_M$ and $ \mathbf{t}^B_M$ represent the extrinsic calibration from the IMU frame $\{M\}$ to the IMU frame $\{B\}$. The ground-truth laser point $\mathbf{p}^{W,gt}_j$ can be expanded as
\begin{align}
\mathbf{p}^{W,gt}_j =& \mathbf{R}^W_B \text{Exp}(\delta \boldsymbol{\theta}^W_B) (\mathbf{R}^B_M (\mathbf{p}^M_j + \delta\mathbf{p}^M_j ) + \mathbf{t}^B_M) \\
&+ \mathbf{t}^W_B + \delta \mathbf{t}^W_B  \nonumber\\
\approx& \mathbf{p}^W_j + \mathbf{J}_{\mathbf{R}^W_B}\delta \boldsymbol{\theta}^W_B + \mathbf{J}_{\mathbf{p}^M_j} \delta\mathbf{p}^M_j + \mathbf{J}_{\mathbf{t}^W_B}\delta \mathbf{t}^W_B, \nonumber
\end{align}
where
\begin{equation*}
\begin{split}
\mathbf{J}_{\mathbf{R}^W_B} &= -\mathbf{R}^W_B [\mathbf{R}^B_M \mathbf{p}^M_j + \mathbf{t}^B_M]_{\times}, \\
\mathbf{J}_{\mathbf{p}^M_j} &= \mathbf{R}^W_B \mathbf{R}^B_M, \\
\mathbf{J}_{\mathbf{t}^W_B} &= \mathbf{I}.
\end{split}
\end{equation*}
The operator $\text{Exp}(\cdot)$ denotes the exponential map \cite{imu_preint}. The vector $\delta \boldsymbol{\theta}^W_B \sim \mathcal{N}(\mathbf{0}, \boldsymbol{\Sigma}_{\mathbf{R}^W_B})$ and $\delta \mathbf{t}^W_B \sim \mathcal{N}(\mathbf{0},\boldsymbol{\Sigma}_{\mathbf{t}^W_B})$ represent the error state of $\mathbf{R}^W_B$ and $\mathbf{t}^W_B$. Then the noise $\delta \mathbf{p}^W_j$ of the point $\mathbf{p}^W_j$ and its covariance $\boldsymbol{\Sigma}_{\mathbf{p}^W_j}$ is
\begin{align}
\delta \mathbf{p}^W_j &= \mathbf{B}_j
\begin{bmatrix}
\delta \boldsymbol{\theta}^W_B \\
\delta \mathbf{p}^M_j \\ 
\delta \mathbf{t}^W_B
\end{bmatrix} \sim \mathcal{N}(\mathbf{0}, \boldsymbol{\Sigma}_{\mathbf{p}^W_j}), \\
\boldsymbol{\Sigma}_{\mathbf{p}^W_j} &=  \mathbf{B}_j \begin{bmatrix}
\boldsymbol{\Sigma}_{\mathbf{R}^W_B} & \mathbf{0} & \mathbf{0} \\
\mathbf{0} & \boldsymbol{\Sigma}_{\mathbf{p^M_j}} & \mathbf{0} \\
\mathbf{0} & \mathbf{0} & \boldsymbol{\Sigma}_{\mathbf{t}^W_B}
\end{bmatrix} \mathbf{B}_j^\top, \nonumber\\
\mathbf{B}_j &= \begin{bmatrix}
\mathbf{J}_{\mathbf{R}^W_B} & \mathbf{J}_{\mathbf{p}^M_j} & \mathbf{J}_{\mathbf{t}^W_B}
\end{bmatrix}. \nonumber
\end{align}

Finally, following the procedure in \cite{adaptive_voxel}, the uncertainty analysis of a laser point $\mathbf{p}^W_j$ is utilized to update the uncertainty of planar features and to construct the point-to-plane residual $\mathbf{r}^{\mathcal{L}}$ in \textcolor{red}{Eq. \eqref{eq:ada_lio_cost_func}}.

\subsection{Observability Analysis of LM-Calibr}\label{sec:supp_obs_analysis_lm_calibr}

In this subsection, Monte Carlo simulations are conducted to identify the unobservability case of LM-Calibr. Specifically, 40 virtual planes are generated to ensure sufficient observations for LM-Calibr. The ground-truth mounting angles are then sampled at $10^\circ$ intervals across the range $[ -180^\circ, 180^\circ ]$ for the DH parameters $\bar{\theta}$ and $\bar{\phi}$. During calibration, the minimum eigenvalue of the Hessian matrix is recorded as a metric for observability analysis. 

\begin{figure}[!t]
\centering
\includegraphics[width=0.48\textwidth]{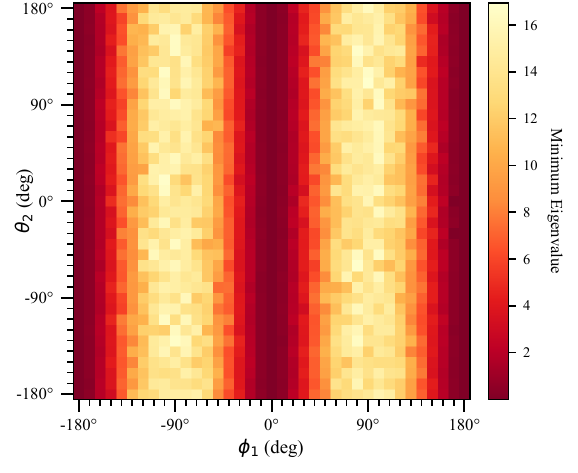}
\caption{LM-Calibr observability analysis for spinning omni LiDAR.}
\label{fig:mid_obs_analysis}
\end{figure}

\begin{figure}[!t]
\centering
\includegraphics[width=0.48\textwidth]{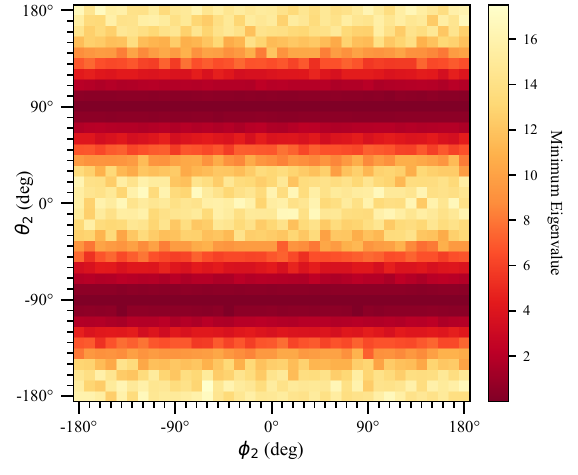}
\caption{LM-Calibr observability analysis for spinning non-omni LiDAR.}
\label{fig:avia_obs_analysis}
\end{figure}

\begin{figure}[!t]
\centering
\includegraphics[width=0.48\textwidth]{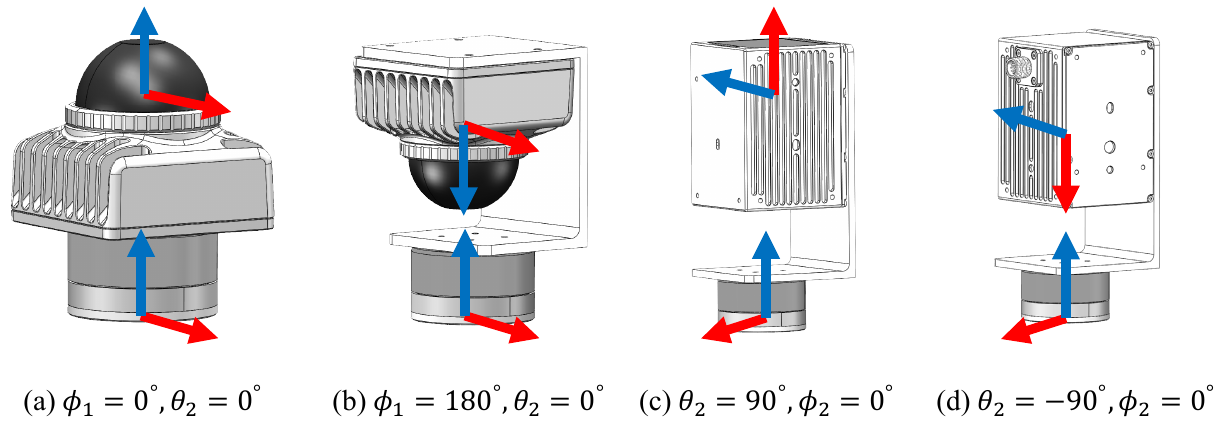}
\caption{The degenerate mounting configurations in LM-Calibr.}
\label{fig:degenerate_mount}
\end{figure}

As illustrated in Fig.~\ref{fig:mid_obs_analysis} and Fig.~\ref{fig:avia_obs_analysis}, a smaller minimum eigenvalue indicates reduced observability of LM-Calibr at the associated mounting angle. For spinning omni LiDAR,
unobservability occurs only when $\phi_1 \approx 0$ or $\pi$, with the corresponding mounting configuration shown in Fig.~\ref{fig:degenerate_mount}(a) and (b). For spinning non-omni LiDAR, LM-Calibr exhibits degeneracy only for the case $\theta_2 \approx \pm \pi/2$, and the associated mounting angle depicted in Fig.~\ref{fig:degenerate_mount}(c) and (d). Note that these degenerate mounting configurations are rarely encountered in practice, as they are physically meaningless for increasing FOV. This analysis confirms that LM-Calibr is applicable to various spinning actuated LiDAR mounting configurations.

\subsection{Identifiability Analysis of LM-Calibr}
\begin{figure}[!t]
\centering
\includegraphics[width=0.48\textwidth]{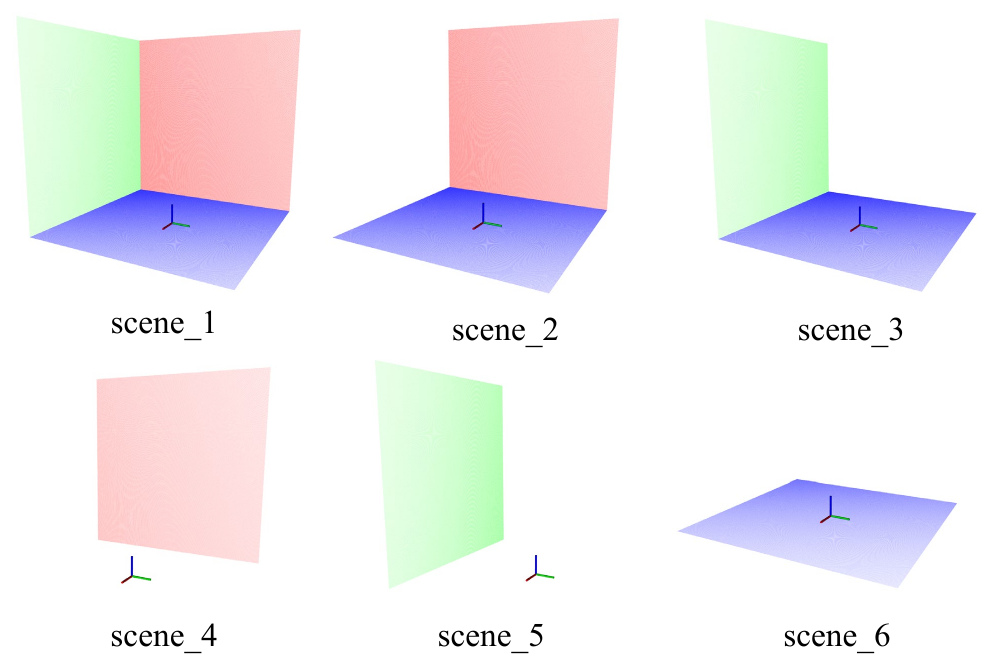}
\caption{The six simulation scenarios in the identifiability analysis of LM-Calibr.}
\label{fig:identifiability}
\end{figure}

This subsection analyzes the relationship between parameter identifiability in LM-Calibr and geometric conditions through six simulation scenarios. As shown in Fig.~\ref{fig:identifiability}, the scenarios include isotropic planar distribution (\texttt{scene\_1}), x-z planar distribution (\texttt{scene\_2}), y-z planar distribution (\texttt{scene\_3}), x planar distribution (\texttt{scene\_4}), y planar distribution (\texttt{scene\_5}), and z planar distribution (\texttt{scene\_6}). Each virtual plane measures $10 m \times 10 m$. The mounting configuration of the spinning Mid360 and spinning Avia are listed in Table \ref{table:id_mount}. The initial values are the ground truth perturbed by $10^\circ$ in rotation and $0.1 m$ in translation.

Tables \ref{table:id_mid_error} and \ref{table:id_avia_error} report calibration errors and eigenvalues for each parameter across varying scenarios. Larger eigenvalues imply enhanced parameter identifiability. The results indicate that translational identifiability degrades significantly in single-plane scenarios. This degradation occurs specifically if the plane's normal vector is parallel to the motor's spinning axis. Practically, the geometric degeneracy in \texttt{scene\_6} can be circumvented by mounting the sensing system laterally during data acquisition. Therefore, LM-Calibr ensures the identifiability of all calibration parameters via only a single plane, e.g., only ground points.

\begin{table}[!t]
\caption{The Mounting Configuration in the Identifiability Analysis of LM-Calibr}\label{table:id_mount}
\setlength{\tabcolsep}{4.3mm}
\centering
\scriptsize
\begin{threeparttable}
\begin{tabular}{ccccc}
\toprule[1pt]
Sensor Type     & $\bar{\theta}$ (deg) & $\bar{d}$ (m)& $\bar{a}$ (m)& $\bar{\phi}$ (deg) \\ \midrule
Spinning Mid360 & -90.0          & 0.5       & 0.1       & 90.0         \\
Spinning Aiva   & 0.0            & 0.1       & 0.5       & 90.0     \\ \bottomrule[1pt]      
\end{tabular}
\end{threeparttable}
\end{table}

\begin{table}[!t]
\caption{The Calibration Errors (mm, deg) and Eigenvalues of LM-Calibr in Spinning Mid360 Calibration}\label{table:id_mid_error}
\setlength{\tabcolsep}{1.0mm}
\centering
\scriptsize
\begin{threeparttable}
\begin{tabular}{c|cc|cc|cc|cc}
\toprule[1pt]
\multirow{2}{*}{Seq.} & \multicolumn{2}{c|}{$\theta_2$} & \multicolumn{2}{c|}{$d_2$} & \multicolumn{2}{c|}{$a_1$} & \multicolumn{2}{c}{$\phi_1$} \\ \cline{2-9} 
                      & Err.         & Eigen.           & Err.        & Eigen.       & Err.        & Eigen.       & Err.        & Eigen.         \\ \hline
scene\_1                    & <0.01         & 63668.53         & 0.55        & 1111.83      & 0.16        & 3448.58      & <0.01        & 24545.62       \\
scene\_2                     & <0.01         & 36318.31         & 1.08        & 623.58       & 0.21        & 1652.05      & <0.01        & 14960.01       \\
scene\_3                     & <0.01         & 38024.62         & 1.38        & 521.95       & 0.56        & 1828.61      & <0.01        & 12956.83       \\
scene\_4                     & <0.01         & 20264.32         & <0.01        & 807.23       & 1.01        & 643.59       & <0.01        & 16668.02       \\
scene\_5                     & <0.01         & 23275.84         & <0.01        & 1035.07      & <0.01        & 552.05       & <0.01        & 14177.53       \\
scene\_6                     & <0.01         & 9934.74          & 53.89       & 0.24         & 584.92      & 0.47         & <0.01        & 2016.19    \\ \bottomrule[1pt]       
\end{tabular}
\end{threeparttable}
\end{table}

\begin{table}[!t]
\caption{The Calibration Errors (mm, deg) and Eigenvalues of LM-Calibr in Spinning Avia Calibration}\label{table:id_avia_error}
\setlength{\tabcolsep}{1.3mm}
\centering
\scriptsize
\begin{threeparttable}
\begin{tabular}{c|cc|cc|cc|cc}
\toprule[1pt]
\multirow{2}{*}{Seq.} & \multicolumn{2}{c|}{$\theta_2$} & \multicolumn{2}{c|}{$d_2$} & \multicolumn{2}{c|}{$a_2$} & \multicolumn{2}{c}{$\phi_2$} \\ \cline{2-9} 
                      & Err.          & Eigen.          & Err.         & Eigen.      & Err.         & Eigen       & Err.        & Eigen.         \\ \hline
scene\_1              & <0.01          & 528.59          & <0.01         & 413.86      & <0.01         & 111.68      & <0.01        & 3584.39        \\
scene\_2              & <0.01          & 283.52          & <0.01         & 296.08      & <0.01         & 81.48       & <0.01        & 2489.28        \\
scene\_3              & <0.01          & 300.77          & <0.01         & 298.53      & <0.01         & 84.53       & <0.01        & 2521.83        \\
scene\_4              & <0.01          & 430.04          & <0.01         & 242.32      & <0.01         & 70.43       & <0.01        & 1334.01        \\
scene\_5              & <0.01          & 484.31          & <0.01         & 376.82      & <0.01         & 109.05      & <0.01        & 1505.41        \\
scene\_6              & 0.02          & 20.48           & 185.05       & <0.01        & 173.46       & 0.02        & <0.01        & 1153.99  \\  \bottomrule[1pt]     
\end{tabular}
\end{threeparttable}
\end{table}

\subsection{The Accuracy of EVA-LIO (w/o spin.) in MCD Benchmark}
EVA-LIO  (w/o spin.) and the baseline methods are evaluated on a representative subset of the MCD Benchmark to demonstrate its generalizability. As shown in Table \ref{table:mcd_dataset}, EVA-LIO  (w/o spin.) achieves comparable or superior localization accuracy. This improvement primarily stems from the pre-point uncertainty estimation, as well as from the environmental analysis module, which enables EVA-LIO (w/o spin.) to select appropriate downsampling rates and voxel maps based on the spatial scale.

\begin{table}[!t]
\caption{Absolute Trajectory Error (m) in MCD Benchmark}\label{table:mcd_dataset}
\setlength{\tabcolsep}{3.0mm}
\centering
\scriptsize
\begin{threeparttable}
\begin{tabular}{cccc}
\toprule[1pt]
Methods             & ntu\_day\_01  & kth\_day\_06  & tuhh\_day\_03 \\ \hline
EVA-LIO (w/o spin.) & 1.58          & \textbf{0.39} & \textbf{0.76} \\
Voxel-SLAM          & 1.97          & 0.42          & 0.90          \\
Point-LIO           & 1.72          & 0.41          & 0.99          \\
Fast-LIO2           & \textbf{1.54} & 0.45          & 0.93          \\
Ada-LIO             & 1.61          & 0.58          & 1.07          \\
CTE-MLO             & 3.88          & 1.44          & 0.89          \\
Traj-LIO            & 1.58          & 0.41          & 0.79       \\ \bottomrule[1pt]     
\end{tabular}
\end{threeparttable}
\end{table}

\subsection{Acceleration Bounds for Stable EVA-LIO Operation}

EVA-LIO relies on IMU measurements to compensate for point cloud motion distortion and to provide an initial estimate for point-to-plane registration. For stable operation, the deviation between the initial estimate and the ground truth motion must be bounded by the point-to-plane correspondence threshold $\varepsilon_p$. Let $T_{scan}$ denote the LiDAR scanning period and $a$ the sensing system acceleration. Under a constant-acceleration approximation, the intra-scan displacement is on the order of 
\begin{equation}
||\Delta p|| = \frac{1}{2} ||a|| T_{scan}^2.
\end{equation}

EVA-LIO operates stably within the bound $||\Delta p|| \leq \varepsilon_p$, which equivalently requires the IMU translation integration noise,  $\sigma_{imu}$, satisfies $\sigma_{imu} \sim \mathcal{N}(0,  \varepsilon_p^2)$. This yields an acceleration upper bound
\begin{equation}
||a||\leq \frac{2\varepsilon_p}{T^2_{scan}}.
\end{equation}

For instance, with $T_{scan} = 0.1s$ and $\varepsilon_p = 0.1m$, the corresponding bound is $||a|| \approx 20 m/s^2$, i.e., around $2g$. According to the $3\sigma$ criterion, if the acceleration reaches $6g$ or above, the laser points are unable to establish correct planar correspondences, leading to a degradation of EVA-LIO performance.  Moreover, due to the limited IMU sampling frequency (about $400 Hz$), the assumptions underlying midpoint integration are violated under such aggressive maneuvers ($\ge 6g$). As a result, motion compensation is significantly compromised by IMU noise and bias errors, which induces pose drift and intra-frame distortion.

\begin{table}[!t]
\caption{The System Parameters of Baseline LIOs in CA and CR}\label{table:param_config_1}
\setlength{\tabcolsep}{4.3mm}
\centering
\scriptsize
\begin{threeparttable}
\begin{tabular}{ccc}
\toprule[1pt]
Methods     & Downsample Rate (m) & Map Resolution (m) \\ \hline
Voxel-SLAM & 0.1                 & 0.5                \\
Point-LIO  & 0.1                 & 0.1                \\
Fast-LIO2  & 0.1                 & 0.1                \\
Ada-LIO    & 0.1                 & 0.2 / 0.5 / 1.2    \\
CTE-MLO    & 0.1                 & 1.0                \\
Traj-LO    & 0.1                 & 0.5         \\    \bottomrule[1pt]   
\end{tabular}
\end{threeparttable}
\end{table}

\begin{table}[!t]
\caption{The System Parameters of Baseline LIOs in BG, NTU, KTH, TUHH and Real-world}\label{table:param_config_2}
\setlength{\tabcolsep}{4.3mm}
\centering
\scriptsize
\begin{threeparttable}
\begin{tabular}{ccc}
\toprule[1pt]
Methods     & Downsample Rate (m) & Map Resolution (m) \\ \hline
Voxel-SLAM & 0.25                & 1.0                \\
Point-LIO  & 0.2                 & 0.25               \\
Fast-LIO2  & 0.2                 & 0.25               \\
Ada-LIO    & 0.2                 & 0.2 / 0.5 / 1.2    \\
CTE-MLO    & 0.2                 & 1.5                \\
Traj-LO    & 0.5                 & 0.5      \\ \bottomrule[1pt]          
\end{tabular}
\end{threeparttable}
\end{table}

\subsection{System Parameters of Baseline LIOs}

To ensure the localization accuracy, specific system parameters of the baseline LIOs are adjusted separately for each sequence. The parameters are listed in Tables \ref{table:param_config_1} and \ref{table:param_config_2}.

\subsection{Future works}
Future work will extend the current ground-centric evaluation to airborne UAV degeneracy scenes \cite{wang2025uavscenes} characterized by weak geometry, rapid altitude and viewpoint change, and aggressive motion, and will explore RL-based active scanning control \cite{li2026aeos} together with transformer-based geometry grounding \cite{deng2025reloc} for failure-aware relocalization under these conditions.

\bibliographystyle{IEEEtran}
\bibliography{IEEEabrv,reference}

\end{document}